\documentclass{article}
\PassOptionsToPackage{numbers}{natbib}
\usepackage[final]{nips_2018}

\usepackage[utf8]{inputenc} 
\usepackage[T1]{fontenc}    

\usepackage{xcolor}
\usepackage{hyperref}       
\hypersetup{
    colorlinks,
    anchorcolor={blue!50!black},
    linkcolor={blue!50!black},
    citecolor={blue!50!black},
    urlcolor={blue}
}

\usepackage{url}            
\usepackage{booktabs}       
\usepackage{amsfonts}       
\usepackage{nicefrac}       
\usepackage{microtype}      

\usepackage{algorithm,algorithmic,amsmath,amssymb}
\usepackage{subcaption,graphicx,enumitem,booktabs}

\newcommand{\E}{\mathbb{E}}
\newcommand{\R}{\mathbb{R}}
\DeclareMathOperator{\Tr}{Tr}
\DeclareMathOperator{\argmin}{\arg\min}
\newcommand{\pdata}{p_{d}}
\newcommand{\pmodel}{p_{g}}
\newcommand{\ld}{\mathcal{L}_\textsc{d}}
\renewcommand{\lg}{\mathcal{L}_\textsc{G}}
\newcommand{\X}{\mathcal{X}}
\newcommand{\rdd}{\R^{d \times d}}

\makeatletter
\usepackage{footmisc}
\DefineFNsymbols{mySymbols}{{\ensuremath\star}{\ensuremath\ddagger}\S\P
   *{**}{\ensuremath{\dagger\dagger}}{\ensuremath{\ddagger\ddagger}}}
\setfnsymbol{mySymbols}

\makeatother

\title{Are GANs Created Equal? A Large-Scale Study}
\author{
  Mario Lucic\thanks{Indicates equal authorship. Correspondence to \texttt{\{lucic,kkurach\}@google.com.}} \\ \And
  Karol Kurach$^\star$ \\ \And
  Marcin Michalski \\ Google Brain \\ \And
  Olivier Bousquet \\ \And
  Sylvain Gelly
}

\begin{document}
\maketitle
\vspace{-5mm}
\begin{abstract}
  Generative adversarial networks (GAN) are a powerful subclass of
generative models. Despite a very rich research activity leading to
numerous interesting GAN algorithms, it is still very hard to assess
which algorithm(s) perform better than others.  We conduct a neutral,
multi-faceted large-scale empirical study on state-of-the art models
and evaluation measures. We find that most models can reach similar
scores with enough hyperparameter optimization and random restarts.
This suggests that improvements can arise from a higher computational
budget and tuning more than fundamental algorithmic changes.  To
overcome some limitations of the current metrics, we also propose
several data sets on which precision and recall can be computed.  Our
experimental results suggest that future GAN research should be based
on more systematic and objective evaluation procedures. Finally, we
did not find evidence that any of the tested algorithms consistently
outperforms the non-saturating GAN introduced in \cite{goodfellow2014generative}.
\end{abstract}

\section{Introduction}\label{sec:introduction}
Generative adversarial networks (GAN) are a powerful subclass of
generative models and were successfully applied to image generation
and editing, semi-supervised learning, and domain adaptation
\cite{radford2015unsupervised,zhang2016stackgan}.  In the GAN
framework the model learns a deterministic transformation $G$ of a
simple distribution $p_z$, with the goal of matching the data
distribution $\pdata$.  This learning problem may be viewed as a
two-player game between the \emph{generator}, which learns how to
generate samples which resemble real data, and a
\emph{discriminator}, which learns how to discriminate between real and
\emph{fake} data. Both players aim to minimize their own cost and the
solution to the game is the Nash equilibrium where neither player can
improve their cost unilaterally \cite{goodfellow2014generative}.

Various flavors of GANs have been recently proposed, both purely
unsupervised
\cite{goodfellow2014generative,arjovsky2017wasserstein,
  gulrajani2017improved,berthelot2017began}
as well as conditional
\cite{mirza2014conditional,odena2016conditional}. While these models
achieve compelling results in specific domains, there is still no
clear consensus on which GAN algorithm(s) perform objectively better
than others.  This is partially due to the lack of robust and
consistent metric, as well as limited comparisons which put all
algorithms on equal footage, including the computational budget to
search over all hyperparameters. Why is it important? Firstly, to help the
practitioner choose a better algorithm from a very large
set. Secondly, to make progress towards better algorithms and their
understanding, it is useful to clearly assess which modifications are
critical, and which ones are only good on paper, but do not make a
significant difference in practice.

The main issue with evaluation stems from the fact that one cannot
explicitly compute the probability $\pmodel(x)$. As a result, classic
measures, such as log-likelihood on the test set, cannot be
evaluated. Consequently, many researchers focused on qualitative
comparison, such as comparing the visual quality of
samples. Unfortunately, such approaches are subjective and possibly
misleading \cite{gerhard2013sensitive}. As a remedy, two evaluation
metrics were proposed to quantitatively assess the performance of
GANs. Both assume access to a pre-trained classifier. \emph{Inception
Score (IS)} \cite{salimans2016improved} is based on the fact that a
good model should generate samples for which, when evaluated by the
classifier, the class distribution has low entropy. At the same time,
it should produce diverse samples covering all classes. In contrast,
\emph{Fr\'echet Inception Distance} is computed by considering the
difference in embedding of true and fake data \cite{heusel2017gans}.
Assuming that the coding layer follows a multivariate Gaussian
distribution, the distance between the distributions is reduced to the
Fr\'echet distance between the corresponding Gaussians.

\textbf{Our main contributions:}
(1) We provide a fair and comprehensive comparison of the
state-of-the-art GANs, and empirically demonstrate that nearly all of
them can reach similar values of FID, given a high enough
computational budget. (2) We provide strong empirical
evidence\footnote{Reproducing these experiments requires approximately
  6.85 GPU years (NVIDIA P100).} that to
compare GANs it is necessary to report a summary of distribution of
results, rather than the best result achieved, due to the randomness
of the optimization process and model instability.  (3) We assess the
robustness of FID to mode dropping, use of a different encoding
network, and provide estimates of the best FID achievable on classic
data sets.  (4) We introduce a series of tasks of increasing
difficulty for which undisputed measures, such as precision and
recall, can be approximately computed.  (5) We open-sourced our
experimental setup and model implementations at
\href{https://github.com/google/compare_gan}{goo.gl/G8kf5J}.

\section{Background and Related Work}\label{sec:related}
There are several ongoing challenges in the study of GANs, including
their convergence and generalization properties
\cite{arora2017generalization, mescheder2017numerics}, and
optimization stability \cite{ salimans2016improved,
arjovsky2017wasserstein}. Arguably, the most critical challenge is
their \emph{quantitative} evaluation.  The classic approach towards
evaluating generative models is based on model likelihood which is
often intractable. While the log-likelihood can be approximated for
distributions on low-dimensional vectors, in the context of complex
high-dimensional data the task becomes extremely
challenging. \citet{wu2016quantitative} suggest an annealed importance
sampling algorithm to estimate the hold-out log-likelihood.  The key
drawback of the proposed approach is the assumption of the Gaussian
observation model which carries over all issues of kernel density
estimation in high-dimensional spaces. \citet{theis2015note} provide
an analysis of common failure modes and demonstrate that it is
possible to achieve high likelihood, but low visual quality, and
vice-versa. Furthermore, they argue against using Parzen window
density estimates as the likelihood estimate is often incorrect. In
addition, ranking models based on these estimates is discouraged
\citep{bachman2015variational}. For a discussion on other drawbacks of
likelihood-based training and evaluation consult
\citet{huszar2015not}.

\textbf{Inception Score (IS)}. Proposed by
\citet{salimans2016improved}, IS offers a way to quantitatively
evaluate the quality of generated samples. The score was motivated by
the following considerations: (i) The conditional label distribution
of samples containing meaningful objects should have low entropy, and
(ii) The variability of the samples should be high, or equivalently,
the marginal $\int_z p(y | x=G(z))dz$ should have high
entropy. Finally, these desiderata are combined into one score,
$\texttt{IS}(G) = \exp(\E_{x \sim G}[d_{KL}(p(y \mid x), p(y)]).$
The classifier is Inception Net trained on Image Net.
The authors found that this score is well-correlated with
scores from human annotators \cite{salimans2016improved}. Drawbacks
include insensitivity to the prior distribution over labels and not
being a proper \emph{distance}.

\textbf{Fr\'echet Inception Distance (FID).} Proposed by
\citet{heusel2017gans}, FID provides an alternative approach. To quantify
the quality of generated samples, they are first embedded into a
feature space given by (a specific layer) of Inception Net. Then,
viewing the embedding layer as a continuous multivariate Gaussian, the
mean and covariance is estimated for both the generated
data and the real data. The Fr\'echet distance between these two
Gaussians is then used to quantify the quality of the samples, i.e.
$\texttt{FID}(x, g) = ||\mu_x - \mu_g||_2^2 + \Tr(\Sigma_x + \Sigma_g -
  2(\Sigma_x\Sigma_g)^\frac12),$
where $(\mu_x, \Sigma_x)$, and $(\mu_g, \Sigma_g)$ are the mean and
covariance of the sample embeddings from the data distribution and
model distribution, respectfully. The authors show that the score is
consistent with human judgment and more robust to noise than IS
\cite{heusel2017gans}. Furthermore, the authors present compelling
results showing negative correlation between the FID and visual
quality of generated samples. Unlike IS, FID can detect intra-class
mode dropping, i.e. a model that generates only one image per class
can score a perfect IS, but will have a bad FID. We provide a thorough
empirical analysis of FID in Section~\ref{sec:metrics}. A significant
drawback of both measures is the inability to detect overfitting. A
``memory GAN'' which stores all training samples would score
perfectly. Finally, as the FID estimator is \emph{consistent},
relative model comparisons for large sample sizes are sound.

\begin{table*}[t]
  \centering
  \scriptsize
  \caption{\small Generator and discriminator loss functions. The main
    difference whether the discriminator outputs a probability (\textsc{mm
      gan}, \textsc{ns gan}, \textsc{dragan}) or its output is unbounded
    (\textsc{wgan}, \textsc{wgan gp}, \textsc{ls gan}, \textsc{began}),
    whether the gradient penalty is present (\textsc{wgan gp},
    \textsc{dragan}) and where is it evaluated.}
  \label{tab:gans}
  \renewcommand{\arraystretch}{1.3}
  \begin{tabular}{lll}
  \toprule
  \textsc{GAN}
  & \textsc{Discriminator Loss}& \textsc{Generator Loss}\\\toprule
  \textsc{mm gan}     
  & $\ld^{\textsc{gan}} = -\E_{x \sim \pdata}[\log(D(x))] - \E_{\hat{x} \sim \pmodel}[\log(1 - D(\hat{x}))]$
  & $\lg^{\textsc{gan}} = \E_{\hat{x} \sim \pmodel}[\log(1 - D(\hat{x}))]$ \\\midrule
  \textsc{ns gan}  
  & $\ld^{\textsc{nsgan}} = -\E_{x \sim \pdata}[\log(D(x))] - \E_{\hat{x} \sim \pmodel}[\log(1 - D(\hat{x}))]$
  & $\lg^{\textsc{nsgan}} = -\E_{\hat{x} \sim \pmodel}[\log(D(\hat{x}))]$ \\\midrule
  \textsc{wgan}
  & $\ld^{\textsc{wgan}} = -\E_{x \sim \pdata}[D(x)] + \E_{\hat{x} \sim \pmodel}[D(\hat{x})]$
  & $\lg^{\textsc{wgan}}= - \E_{\hat{x} \sim \pmodel}[D(\hat{x})]$ \\\midrule
  \textsc{wgan gp}
  & $\ld^{\textsc{wgangp}} =\ld^{\textsc{wgan}} + \lambda\E_{\hat{x} \sim \pmodel}[(||\nabla D(\alpha x + (1-\alpha \hat{x})||_2 - 1)^2]$ 
  & $\lg^{\textsc{wgangp}}=-\E_{\hat{x} \sim \pmodel}[D(\hat{x})]$ \\\midrule
  \textsc{ls gan} 
  & $\ld^{\textsc{lsgan}} = -\E_{x \sim \pdata}[(D(x) - 1)^2] + \E_{\hat{x} \sim \pmodel}[D(\hat{x})^2]$
  & $\lg^{\textsc{lsgan}}=-\E_{\hat{x} \sim \pmodel}[(D(\hat{x} - 1))^2]$ \\\midrule
  \textsc{dragan} 
  & $\ld^{\textsc{dragan}} = \ld^{\textsc{gan}} + \lambda\E_{\hat{x} \sim \pdata + \mathcal{N}(0, c)}[(||\nabla D(\hat{x})||_2 - 1)^2]$
  & $\lg^{\textsc{dragan}} = \E_{\hat{x} \sim \pmodel}[\log(1 - D(\hat{x}))]$ \\\midrule
  \textsc{began} 
  & $\ld^{\textsc{began}} = \E_{x \sim \pdata}[||x - \textsc{AE}(x)||_1] - k_t\E_{\hat{x} \sim \pmodel}[||\hat{x} - \textsc{AE}(\hat{x})||_1]$
  & $\lg^{\textsc{began}} = \E_{\hat{x} \sim \pmodel}[||\hat{x} - \textsc{AE}(\hat{x})||_1]$ \\\midrule
\end{tabular}

  \vspace{-3mm}
\end{table*}

A very recent study comparing several GANs using IS has been presented
by \citet{fedus2017many}. The authors focus on IS and consider a
smaller subset of GANs. In contrast, our focus is on providing a
\emph{fair assessment} of the current state-of-the-art GANs using FID,
as well as precision and recall, and also verifying the robustness of
these models in a large-scale empirical evaluation.

\section{Flavors of Generative Adversarial Networks}\label{sec:gans}
In this work we focus on \emph{unconditional} generative adversarial
networks.  In this setting, only unlabeled data is available for
learning. The optimization problems arising from existing approaches
differ by (i) the constraint on the discriminators output and
corresponding loss, and the presence and application of gradient norm
penalty.

In the original GAN formulation \cite{goodfellow2014generative} two
loss functions were proposed. In the \emph{minimax} GAN the
discriminator outputs a probability and the loss function is the
negative log-likelihood of a binary classification task (\textsc{mm
  gan} in Table~\ref{tab:gans}). Here the generator learns to generate
samples that have a low probability of being fake. To improve the
gradient signal, the authors also propose the \emph{non-saturating}
loss (\textsc{ns gan} in Table~\ref{tab:gans}), where the generator
instead aims to maximize the probability of generated samples
being real. In Wasserstein GAN \cite{arjovsky2017wasserstein} the discriminator is
allowed to output a real number and the objective function is
equivalent to the MM GAN loss without the sigmoid (WGAN in
Table~\ref{tab:gans}). The authors prove that, under an optimal
(Lipschitz smooth) discriminator, minimizing the value function with
respect to the generator minimizes the Wasserstein distance between
model and data distributions. Weights of the discriminator are clipped
to a small absolute value to enforce smoothness. To improve on the
stability of the training, \citet{gulrajani2017improved} instead add a
soft constraint on the norm of the gradient which encourages the discriminator
to be 1-Lipschitz. The gradient norm is evaluated on points obtained by
linear interpolation between data points and generated samples where
the optimal discriminator should have unit gradient norm
\citep{gulrajani2017improved}. Gradient norm penalty can also be added to both \textsc{mm gan} and \textsc{ns gan}
and evaluated around the data manifold (\textsc{dragan} \cite{kodali2017dragan} in
Table~\ref{tab:gans} based on \textsc{ns gan}). This encourages the
discriminator to be piecewise linear around the data manifold.  Note
that the gradient norm can also be evaluated between fake and real
points, similarly to \textsc{wgan gp}, and added to either \textsc{mm gan} or
\textsc{ns gan} \cite{fedus2017many}. \citet{mao2016least} propose a least-squares loss for the
discriminator and show that minimizing the corresponding objective
(\textsc{ls gan} in Table~\ref{tab:gans}) implicitly minimizes the
Pearson $\chi^2$ divergence. The idea is to provide smooth loss which
saturates slower than the sigmoid cross-entropy loss of the original
\textsc{mm gan}. Finally, \citet{berthelot2017began} propose to use an autoencoder
as a discriminator and optimize a lower bound of the Wasserstein distance
between auto-encoder \emph{loss distributions} on real and fake
data. They introduce an additional hyperparameter $\gamma$ to control
the equilibrium between the generator and discriminator.

\section{Challenges of a Fair Comparison}\label{sec:comparison}

There are several interesting dimensions to this problem, and there is
no single \emph{right way} to compare these models (i.e. the loss
function used in each GAN). Unfortunately, due to the combinatorial
explosion in the number of choices and their ordering, not all
relevant options can be explored. While there is \emph{no definite answer} on
how to best compare two models, in this work we have made several
pragmatic choices which were motivated by two practical concerns: providing
a neutral and fair comparison, and a hard limit on the computational budget.

\textbf{Which metric to use?} Comparing models implies access to some
metric. As discussed in Section~\ref{sec:related}, classic measures,
such as model likelihood cannot be applied. We will argue for and
study two sets of evaluation metrics in Section~\ref{sec:metrics}: FID, which
can be computed on all data sets, and precision, recall, and $F_1$, which we
can compute for the proposed tasks.

\textbf{How to compare models?}
Even when the metric is fixed, a given algorithm can achieve very
different scores, when varying the architecture, hyperparameters,
random initialization (i.e. random seed for initial network weights),
or the data set. Sensible targets include best score across all
dimensions (e.g. to claim the best performance on a fixed data set),
average or median score (rewarding models which are good in
expectation), or even the worst score (rewarding models with
worst-case robustness). These choices can even be combined --- for
example, one might train the model multiple times using the best
hyperparameters, and average the score over random initializations).

For each of these dimensions, we took several pragmatic choices to
reduce the number of possible configurations, while still exploring
the most relevant options.
\begin{enumerate}[topsep=0pt,itemsep=1pt,partopsep=0pt,parsep=0pt,leftmargin=5mm]
\item \textbf{Architecture}: We use the \emph{same} architecture for
  all models. We note that this architecture suffices to achieve good performance on considered data sets.
\item \textbf{Hyperparameters}: For both training hyperparameters
  (e.g.  the learning rate), as well as model specific ones
  (e.g. gradient penalty multiplier), there are two valid approaches:
  (i) perform the hyperparameter optimization for each data set, or (ii)
  perform the hyperparameter optimization on one data set and
  \emph{infer} a good range of hyperparameters to use on other data
  sets. We explore both avenues in
  Section~\ref{sec:fidexperiments}. \item \textbf{Random seed}: Even
  with everything else being fixed, varying the random seed may influence on the results.
  We study this effect and report the corresponding confidence intervals.
\item \textbf{Data set}: We chose four popular data sets from GAN literature.
\item \textbf{Computational budget}: Depending on the budget to
  optimize the parameters, different algorithms can achieve the best
  results. We explore how the results vary depending on the budget $k$,
  where $k$ is the number of hyperparameter settings for a fixed model.
  
\end{enumerate}

In practice, one can either use hyperparameter values suggested by
respective authors, or try to optimize
them. Figure~\ref{fig:phase1_boxplot} and in particular
Figure~\ref{fig:phase2_boxplot} show that optimization is
necessary. Hence, we optimize the hyperparameters for each model and
data set by performing a random search. While we present the results
which were obtained by a random search, we have also investigated
sequential Bayesian optimization, which resulted in comparable
results. We concur that the models with fewer hyperparameters have an
advantage over models with many hyperparameters, but consider this
fair as it reflects the experience of practitioners searching for good
hyperparameters for their setting.

\section{Metrics}\label{sec:metrics}
\begin{figure}[t]
  \centering
  \begin{subtable}[b]{0.30\textwidth}
    \begin{tabular}{lrr}
\toprule
      \textsc{Data set} &  \textsc{Avg. FID} &  \textsc{Dev. FID} \\
\midrule
        \textsc{celeba} &      2.27 &      0.02 \\
       \textsc{cifar10} &      5.19 &      0.02 \\
 \textsc{fashion-mnist} &      2.60 &      0.03 \\
        \textsc{mnist} &      1.25 &      0.02 \\
\bottomrule
\end{tabular}

    \subcaption{\scriptsize Bias and variance}
    \label{tab:fid_split_sensitivity}
  \end{subtable}
  \qquad
  \begin{subfigure}[b]{0.30\textwidth}
    \includegraphics[width=\textwidth]{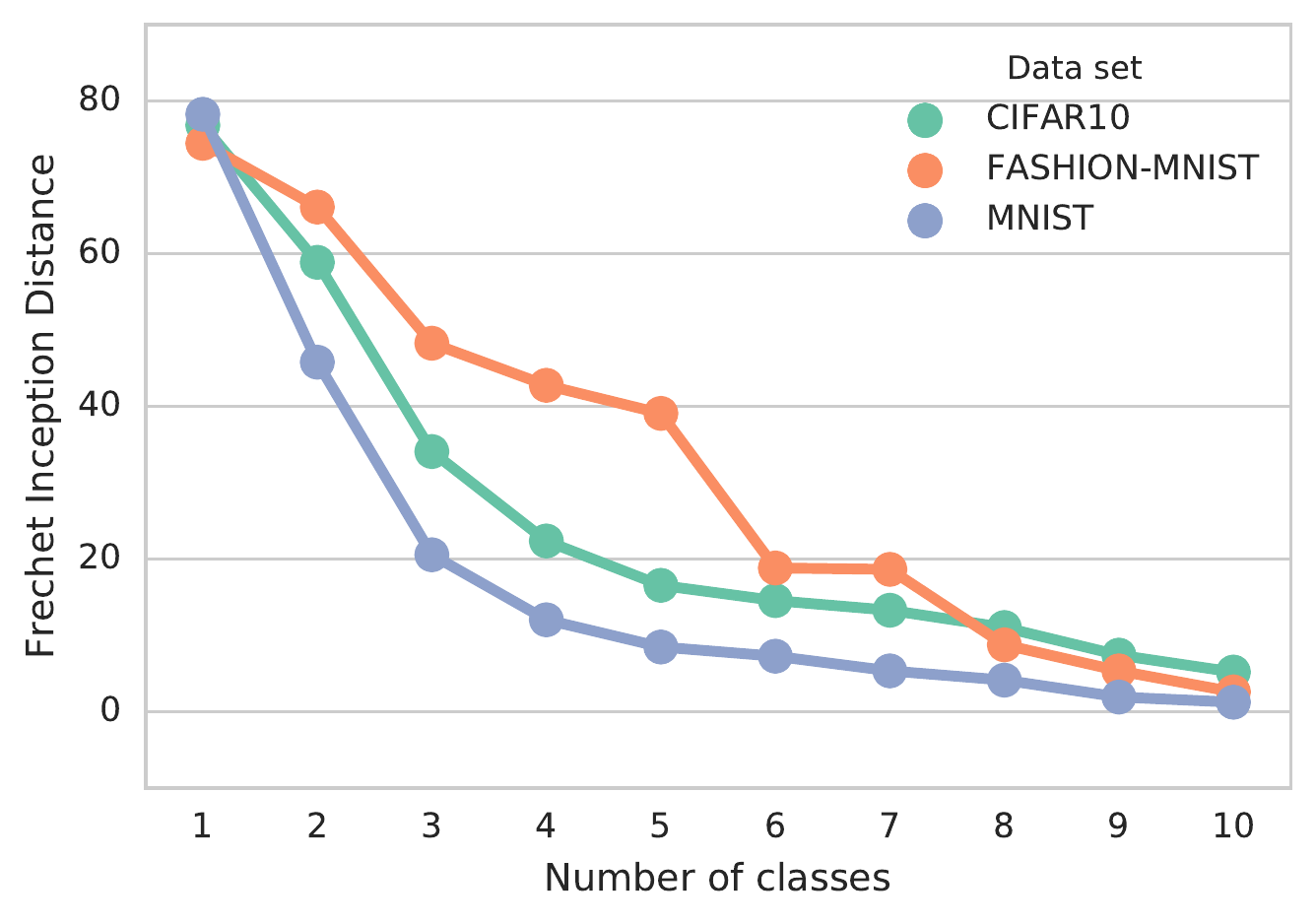}
    \subcaption{\scriptsize Mode dropping}
  \end{subfigure}
  \qquad
  \begin{subfigure}[b]{0.25\textwidth}
    \includegraphics[width=\columnwidth]{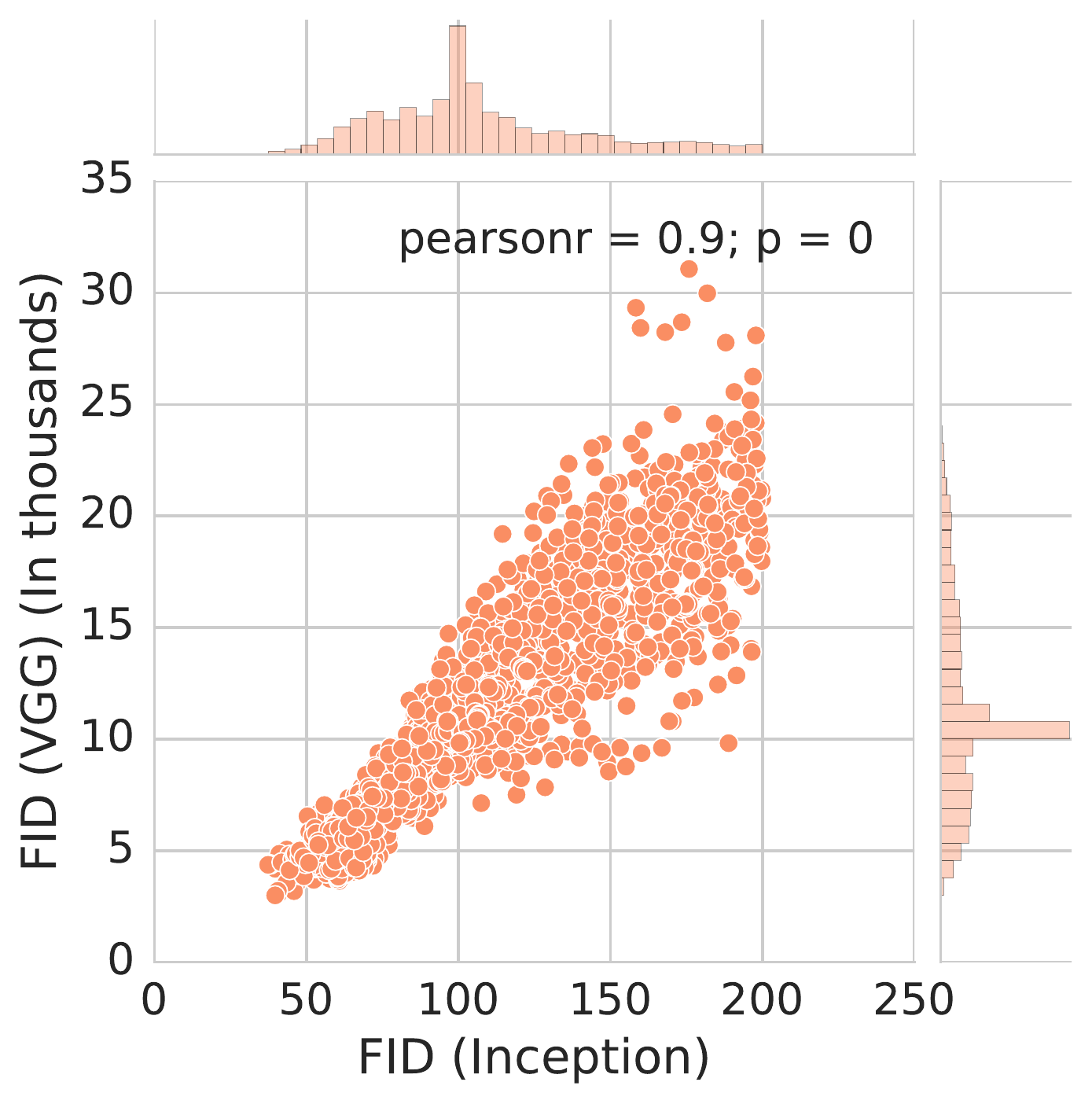}
    \subcaption{\scriptsize VGG vs Inception}
  \end{subfigure}
  \caption{\small Figure (a) shows that FID has a slight bias, but low
    variance on samples of size 10000. Figure (b) shows that FID is
    extremely sensitive to mode dropping. Figure (c) shows the high rank
    correlation (Spearman's $\rho=0.9$) between FID score computed on
    InceptionNet vs FID computed using VGG for the \textsc{CelebA} data
    set (for interesting range: FID~$ < 200$).
    \label{fig:fid_investigation}}
  \vspace{-5mm}
\end{figure}
In this work we focus on two sets of metrics. We first analyze the recently
proposed FID in terms of robustness (of the metric itself), and
conclude that it has desirable properties and can be used in
practice. Nevertheless, this metric, as well as Inception Score, is
incapable of detecting overfitting: a \emph{memory GAN} which simply
stores all training samples would score perfectly under both
measures. Based on these shortcomings, we propose an approximation to
precision and recall for GANs and how that it can be used to
quantify the degree of overfitting. We stress that the proposed method
should be viewed as complementary to IS or FID, rather than a replacement.

\textbf{Fr\'echet Inception Distance.}
FID was shown to be robust to noise~\cite{heusel2017gans}. Here we
quantify the bias and variance of FID, its sensitivity to the encoding
network and sensitivity to mode dropping. To this end, we partition
the data set into two groups, i.e. $\X = \X_1 \cup \X_2$. Then, we
define the data distribution $\pdata$ as the empirical distribution on
a random subsample of $\X_1$ and the model distribution $\pmodel$ to
be the empirical distribution on a random subsample from
$\X_2$. For a random partition this ``model
distribution'' should follow the data distribution.

\begin{figure}[!t]
  \begin{centering}
    \begin{subfigure}[t]{1.35in} \centering
      \includegraphics[width=1.35in]{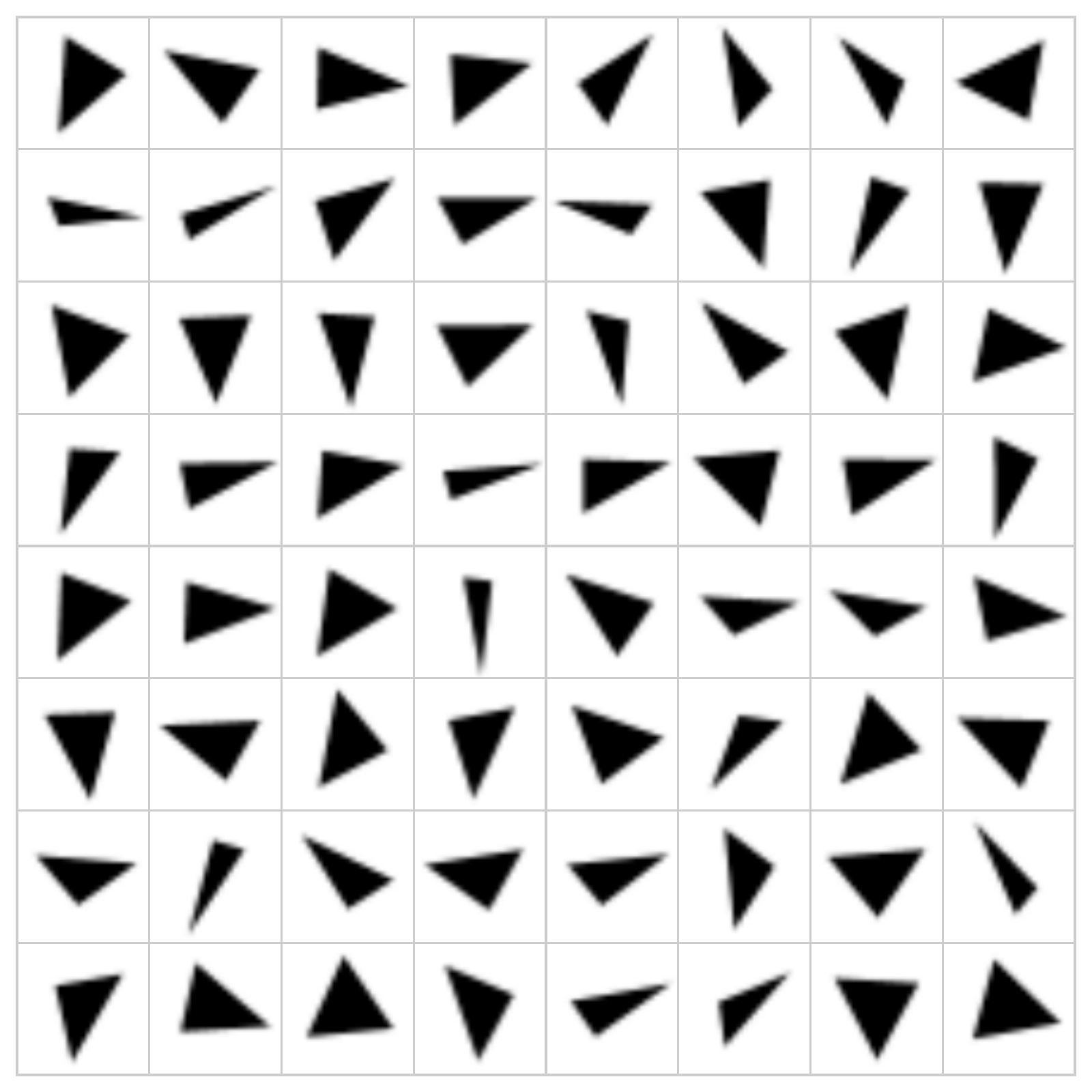}
      \caption{\scriptsize High precision, high recall}\label{fig:distance_to_manifold_a}
    \end{subfigure}
    \begin{subfigure}[t]{1.35in} \centering
      \includegraphics[width=1.35in]{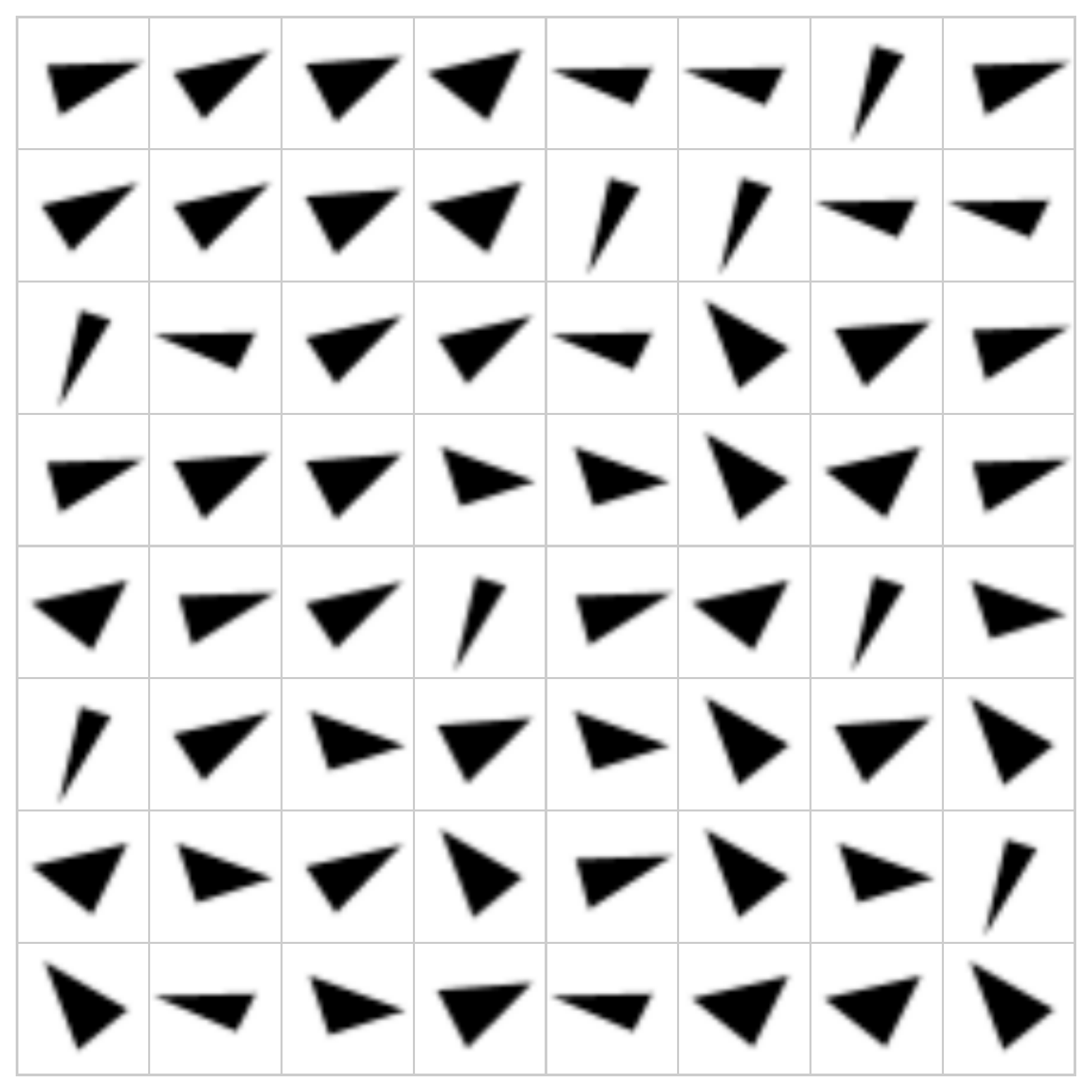}
      \caption{\scriptsize High precision, low recall}
      \label{fig:triangles_high_precision_high_recall}
    \end{subfigure}
    \begin{subfigure}[t]{1.35in} \centering
      \includegraphics[width=1.35in]{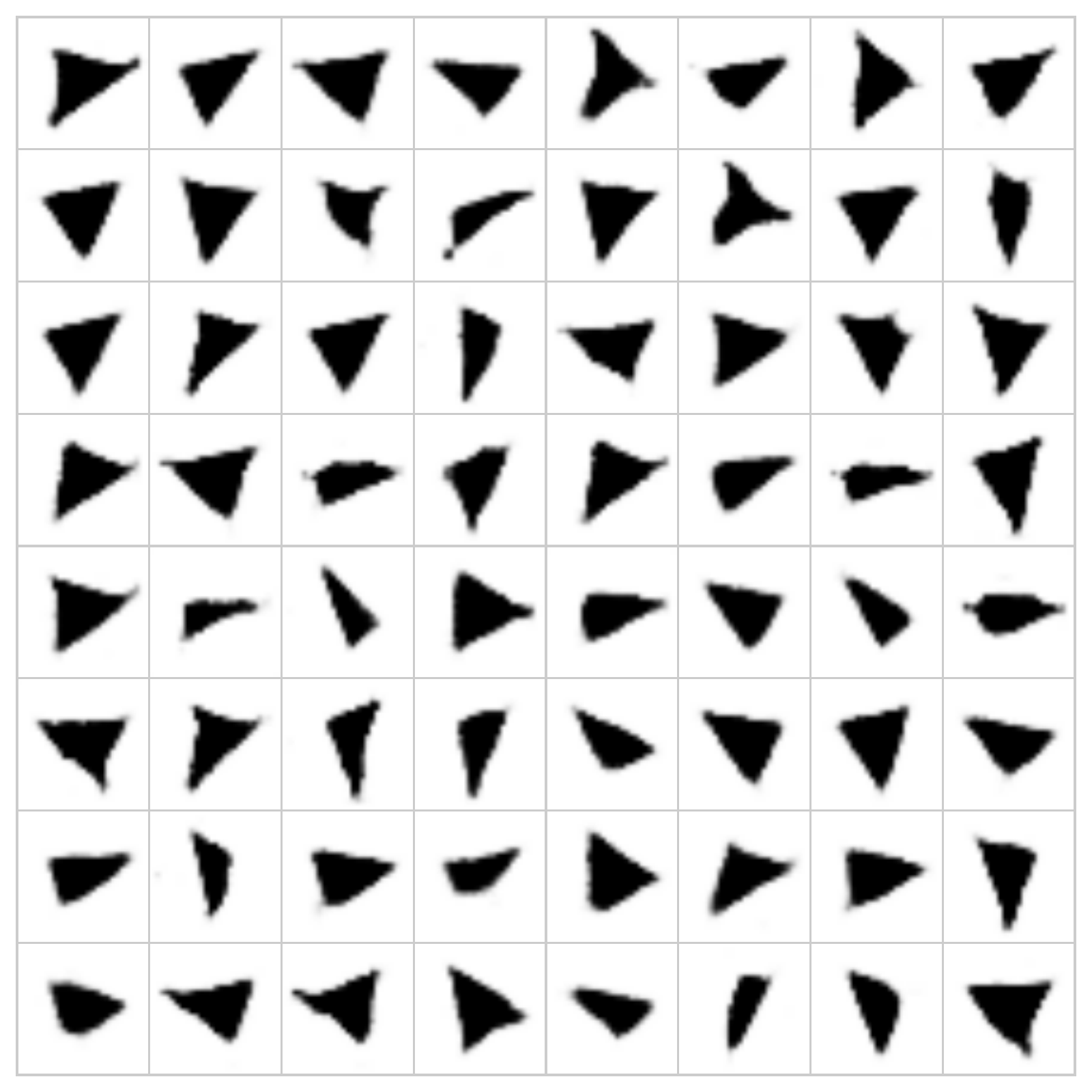}
      \caption{\scriptsize Low precision, high recall}
      \label{fig:triangles_low_precision}
    \end{subfigure}
    \begin{subfigure}[t]{1.35in} \centering
      \includegraphics[width=1.35in]{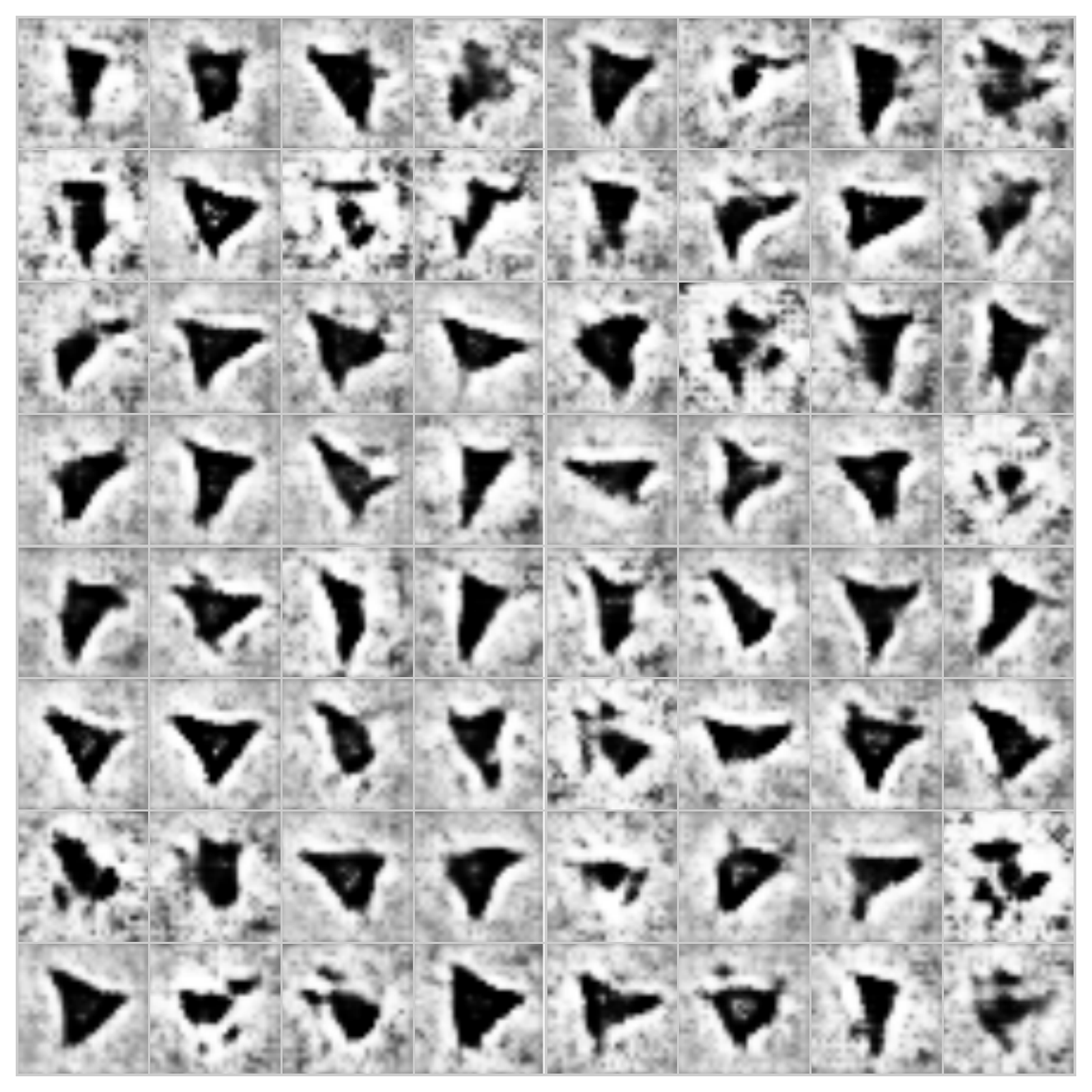}
      \caption{\scriptsize Low precision, low
        recall}\label{fig:hi_precision_low_recall}
    \end{subfigure}
    \caption{\small Samples from models with (a) high
      recall and precision, (b) high precision, but low recall (lacking in
      diversity), (c) low precision, but high recall (can decently reproduce
      triangles, but fails to capture convexity), and (d)
      low precision and low recall.}
    \label{fig:distance_to_manifold}
    \vspace{-5mm}
  \end{centering}
\end{figure}

We evaluate the bias and variance of FID on four data sets
from the GAN literature. We start by using the default train
vs.~test partition and compute the FID between the test set (limited
to $N=10000$ samples for CelebA) and a sample of size $N$ from the
train set. Sampling from the train set is performed $M = 50$
times. The optimistic estimates of FID are reported in
Table~\ref{fig:fid_investigation}. We observe that FID has high
bias, but small variance. From this perspective, estimating the full
covariance matrix might be unnecessary and counter-productive, and a
constrained version might suffice. To test the sensitivity to 
train vs.~test partitioning, we consider $50$ random
partitions (keeping the relative sizes fixed, i.e. $6:1$ for MNIST)
and compute the FID with $M=1$ sample. We observe results similar to
Table~\ref{fig:fid_investigation} which is expected as both training
and testing data sets are sampled from the same distribution. Furthermore, we
evaluate the sensitivity to mode dropping as follows: we fix a
partition $\X = \X_1 \cup \X_2$ and subsample $\X_2$ while keeping only samples
from the first $k$ classes, increasing $k$ from
$1$ to $10$. For each $k$, we consider $50$ random subsamples from
$\X_2$.  Figure~\ref{fig:fid_investigation} shows that FID is heavily
influenced by the missing modes. Finally, we estimate the sensitivity
to the choice of the encoding network by computing FID using the
4096 dimensional \textsc{fc7} layer of the VGG network trained on ImageNet.
Figure~\ref{fig:fid_investigation} shows the resulting distribution. We observe high Spearman's rank correlation
($\rho=0.9$) which encourages the use of the coding layer suggested by the authors.

\textbf{Precision, recall and F$_1$ score.}
Precision, recall and $F_1$ score are proven and widely adopted techniques
for quantitatively evaluating the quality of discriminative
models. Precision measures the fraction of relevant retrieved
instances among the retrieved instances, while recall measures the
fraction of the retrieved instances among relevant instances. $F_1$ score is
the harmonic average of precision and recall. Notice that IS mainly
captures precision: It will not penalize the model for not producing
all modes of the data distribution --- it will only penalize the model
for not producing all \emph{classes}.  On the other hand, FID captures
both precision and recall. Indeed, a model which fails to recover
different modes of the data distribution will suffer in terms of FID.

We propose a simple and effective data set for evaluating (and
comparing) generative models.  Our main motivation is that the
currently used data sets are either too simple (e.g. simple mixtures
of Gaussians, or MNIST) or too complex (e.g. ImageNet). We argue that
it is critical to be able to increase the complexity of the task in a
relatively smooth and controlled fashion.  To this end, we present a
set of tasks for which we can \emph{approximate} the precision and
recall of each model. As a result, we can compare different models
based on established metrics.
The main idea is to construct a data manifold such that the distances
from samples to the manifold can be computed
efficiently. As a
result, the problem of evaluating the quality of the generative model
is effectively transformed into a problem of computing the distance to
the manifold. This enables an intuitive approach for defining the
quality of the model. Namely, if the samples from the model
distribution $\pmodel$ are (on average) close to the manifold, its
\emph{precision} is high. Similarly, high \emph{recall} implies that
the generator can recover (i.e. generate something close to) any sample from
the manifold.

For general data sets, this reduction is impractical as one has to
compute the distance to the manifold which we are trying to
learn. However, if we \emph{construct a manifold} such that this
distance is efficiently computable, the precision and recall can be
efficiently evaluated. To this end, we propose a set of toy data sets
for which such computation can be performed efficiently: The manifold
of convex polygons. As the simplest example, let us focus on
gray-scale triangles represented as one channel images as in
Figure~\ref{fig:distance_to_manifold}. These triangles belong to a
low-dimensional manifold $\mathcal{C}_3$ embedded in
$\rdd$. Intuitively, the coordinate system of this manifold represents
the axes of variation (e.g. rotation, translation, minimum angle size,
etc.). A good generative model should be able to capture these factors
of variation and recover the training samples. Furthermore, it should
recover any sample from this manifold from which we can efficiently
sample which is illustrated in Figure~\ref{fig:distance_to_manifold}.

\textbf{Computing the distance to the manifold.}
Let us consider the simplest case: single-channel gray scale images
represented as vectors $x \in \R^{d^2}$. The distance of a sample $\hat{x}
\in \R^{d^2}$ to the manifold is defined as the squared Euclidean distance to
the closest sample from the manifold $\mathcal{C}_3$, i.e.
$\min_{x \in \mathcal{C}_3} \ell(x, \hat{x}) = \sum_{i = 1}^{d^2}
  ||x_i - \hat{x}_i||_2^2.$
This is a non-convex optimization problem. We find an approximate
solution by gradient descent on the vertices of the triangle (more
generally, a convex polygon), ensuring that each iterate is a valid
triangle (more generally, a convex polygon). To reduce the
false-negative rate we repeat the algorithm 5 times from random
initial solutions. To compute the latent representation of a sample
$\hat{x} \in \rdd$ we \emph{invert} the generator, i.e. we solve
$z^\star = \argmin_{z \in \R^{d_z}} ||\hat{x} - G(z)||_2^2$, using
gradient descent on $z$ while keeping G fixed
\cite{mahendran2015understanding}.

\begin{figure*}[t]
  \centering
  \includegraphics[width=\columnwidth]{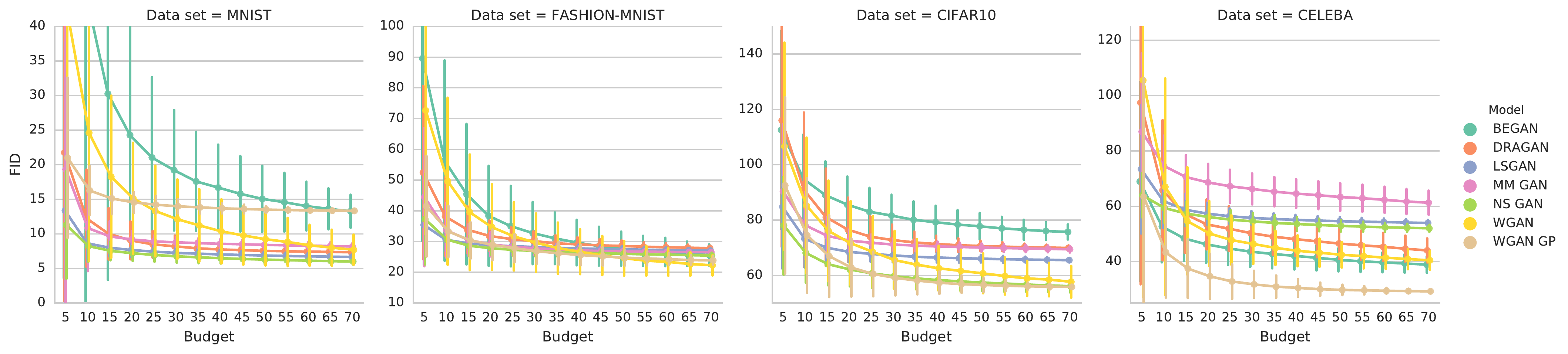}
  \caption{\small How does the minimum FID behave as a function of the
    budget?  The plot shows the distribution of the minimum FID achievable
    for a fixed budget along with one standard deviation interval. For
    each budget, we estimate the mean and variance using $5000$ bootstrap
    resamples out of 100 runs. We observe that, given a relatively low
    budget, all models achieve a similar minimum FID. Furthermore, for a fixed FID, ``bad'' models
    can outperform ``good'' models given enough computational budget. We
    argue that the computational budget to search over hyperparameters is
    an important aspect of the comparison between algorithms.\vspace{-2mm}}
  \label{fig:fid_vs_budget_wide}
\end{figure*}

\section{Large-scale Experimental Evaluation}\label{sec:fidexperiments}

We consider two budget-constrained experimental setups whereby in the
(i) \textbf{wide one-shot setup} one may select $100$ samples of
hyper-parameters per model, and where the range for each
hyperparameter is \emph{wide}, and (ii) the \textbf{narrow two-shots setup} where
one is allowed to select $50$ samples from more narrow ranges which
were manually selected by first performing the wide hyperparameter search
over a specific data set. For the exact ranges and hyperparameter
search details we refer the reader to the
Appendix~\ref{app:hyperparameters}. In the second set of experiments
we evaluate the models based on the "novel" metric: $F_1$ score on the
proposed data set. Finally, we included the Variational Autoencoder
\cite{kingma2013auto} in the experiments as a popular alternative.

\textbf{Experimental setup.}
To ensure a fair comparison, we made the following choices: (i) we use
the generator and discriminator architecture from \textsc{info gan}
\cite{chen2016infogan} as the resulting function space is rich enough
and all considered GANs were not originally designed for this
architecture. Furthermore, it is similar to a proven architecture used
in \textsc{dcgan} \cite{radford2015unsupervised}. The exception is
\textsc{began} where an autoencoder is used as the discriminator.  We
maintain similar expressive power to \textsc{info gan} by using
identical convolutional layers the encoder and approximately matching
the total number of parameters.

For all experiments we fix the latent code size to $64$ and the prior
distribution over the latent space to be uniform on $[-1, 1]^{64}$, except for
\textsc{vae} where it is Gaussian $\mathcal{N}(0, \textbf{I})$. We choose Adam
\cite{kingma2014adam} as the optimization algorithm as it
was the most popular choice in the GAN literature (cf. Appendix~\ref{app:solver} for an empirical comparison to RMSProp).
We apply the same learning rate for both generator and discriminator. We set the batch size
to $64$ and perform optimization for $20$ epochs on \textsc{mnist} and
\textsc{fashion mnist}, 40 on \textsc{CelebA} and 100 on \textsc{cifar}. These data sets
are a popular choice for generative modeling, range from simple to medium complexity,
which makes it possible to run many experiments as well as getting decent results.

Finally, we allow for recent suggestions, such as batch normalization
in the discriminator, and imbalanced update frequencies of generator
and discriminator. We explore these possibilities, together with
learning rate, parameter $\beta_1$ for \textsc{adam}, and
hyperparameters of each model.  We report the hyperparameter ranges
and other details in Appendix~\ref{app:hyperparameters}.

\begin{figure*}[t]
  \centering
  \includegraphics[width=\textwidth]{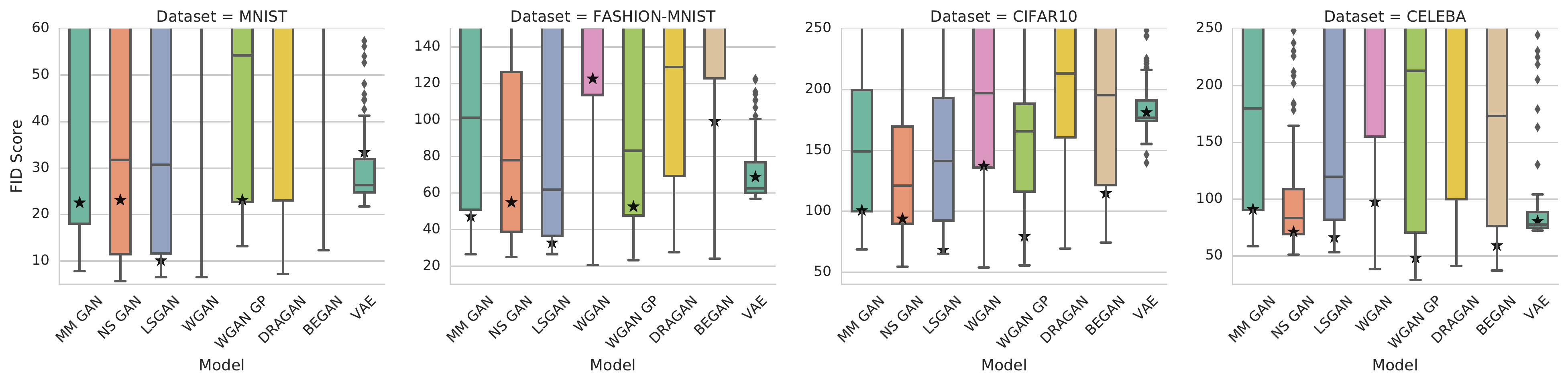}
  \caption{\small A \emph{wide range} hyperparameter search (100
    hyperparameter samples per model). Black stars indicate the
    performance of suggested hyperparameter settings. We observe that
    GAN training is extremely sensitive to hyperparameter settings and
    there is no model which is significantly more stable than others. \label{fig:phase1_boxplot} \vspace{-5mm}
  }

\end{figure*}

\begin{table}[b]
  \centering
  \vspace{-5mm}
  \caption{\small Best FID obtained in a large-scale hyperparameter
    search for each data set. The asterisk (*) on some combinations of models and data sets
    indicates the presence of significant outlier runs, usually severe mode
    collapses or training failures (** indicates up to 20\% failures).  We
    observe that the performance of each model heavily depends on the data
    set and no model strictly dominates the others. 
    Note that these results are \textbf{not ``state-of-the-art''}: (i) larger architectures could improve all models,
    (ii) authors often report the best FID which opens the door for random seed optimization. \label{tab:best_fid_per_dataset}\\}
  {
    \scriptsize
    
\begin{tabular}{lrrrr}
\toprule
\small
     &  MNIST &  FASHION &  CIFAR &  CELEBA \\
\midrule
\textsc{MM GAN}  & $9.8 \pm 0.9$  & $29.6 \pm 1.6$  & $72.7 \pm 3.6$  & $65.6 \pm 4.2$ \\ 
\textsc{NS GAN}  & $6.8 \pm 0.5$  & $26.5 \pm 1.6$  & $58.5 \pm 1.9$  & $55.0 \pm 3.3$ \\ 
\textsc{LSGAN}  & $7.8 \pm 0.6$*  & $30.7 \pm 2.2$  & $87.1 \pm 47.5$  & $53.9 \pm 2.8$* \\ 
\textsc{WGAN}  & $6.7 \pm 0.4$  & $21.5 \pm 1.6$  & $55.2 \pm 2.3$  & $41.3 \pm 2.0$ \\ 
\textsc{WGAN GP}  & $20.3 \pm 5.0$  & $24.5 \pm 2.1$  & $55.8 \pm 0.9$  & $30.0 \pm 1.0$ \\ 
\textsc{DRAGAN}  & $7.6 \pm 0.4$  & $27.7 \pm 1.2$  & $69.8 \pm 2.0$  & $42.3 \pm 3.0$ \\ 
\textsc{BEGAN}  & $13.1 \pm 1.0$  & $22.9 \pm 0.9$  & $71.4 \pm 1.6$  & $38.9 \pm 0.9$ \\ 
\textsc{VAE}  & $23.8 \pm 0.6$  & $58.7 \pm 1.2$  & $155.7 \pm 11.6$  & $85.7 \pm 3.8$ \\ 
\bottomrule 
\end{tabular}
  }
\end{table}

\textbf{A large hyperparameter search.}
We perform hyperparameter optimization and, for each run, look for the \emph{best}
FID across the training run (simulating early stopping). To choose the \emph{best} model,
every $5$ epochs we compute the FID between the $10$k samples
generated by the model and the $10$k samples from the test set.  We
have performed this computationally expensive search for each data
set. We present the sensitivity of models to the hyper-parameters in
Figure~\ref{fig:phase1_boxplot} and the best FID achieved by each
model in Table~\ref{tab:best_fid_per_dataset}. We compute the best FID,
in two phases: We first run a large-scale search on a wide range of
hyper-parameters, and select the best model. Then, we re-run the
training of the selected model $50$ times with different
initialization seeds, to estimate the stability of the training and
report the mean FID and standard deviation, excluding outliers.

Furthermore, we consider the \emph{mean} FID as the computational
budget increases which is shown in
Figure~\ref{fig:fid_vs_budget_wide}.  There are three important
observations. Firstly, there is no algorithm which clearly dominates
others. Secondly, for an interesting range of FIDs, a ``bad'' model
trained on a large budget can out perform a ``good'' model trained on
a small budget. Finally, when the budget is limited, any statistically
significant comparison of the models is unattainable.

\textbf{Impact of limited computational budget.}
In some cases, the computational budget available to a practitioner is too
small to perform such a
large-scale hyperparameter search. Instead, one can tune the
range of hyperparameters on one data set and interpolate the good
hyperparameter ranges for other data sets. We now consider this
setting in which we allow only $50$ samples from a set of narrow
ranges, which were selected based on the wide hyperparameter search on
the \textsc{fashion-mnist} data set. We report the narrow
hyperparameter ranges in
Appendix~\ref{app:hyperparameters}. Figure~\ref{fig:phase2_boxplot}
shows the variance of FID per model, where the hyperparameters were
selected from narrow ranges.  From the practical point of view, there
are significant differences between the models: in some cases the
hyperparameter ranges \emph{transfer} from one data set to the others
(e.g.  \textsc{ns gan}), while others are more sensitive to this
choice (e.g.  \textsc{wgan}).  We note that better scores can be
obtained by a wider hyperparameter search. These results supports the
conclusion that discussing the \emph{best} score obtained by a model
on a data set is not a meaningful way to discern between these
models. One should instead discuss the distribution of the obtained
scores.

\textbf{Robustness to random initialization.}
For a fixed model, hyperparameters, training algorithm, and the order
that the data is presented to the model, one would expect similar
model performance. To test this hypothesis we re-train the best models
from the limited hyperparameter range considered for the previous
section, while changing the initial weights of the generator and
discriminator networks (i.e. by varying a random
seed). Table~\ref{tab:best_fid_per_dataset} and
Figure~\ref{fig:fid_variance_fixed_hyperparameters} show the results
for each data set. Most models are relatively robust to random
initialization, except \textsc{lsgan}, even though for all of them the
variance is significant and should be taken into account when
comparing models.

\textbf{Precision, recall, and F$_1$.}
\begin{figure*}[t]
  \centering
  \includegraphics[width=\columnwidth]{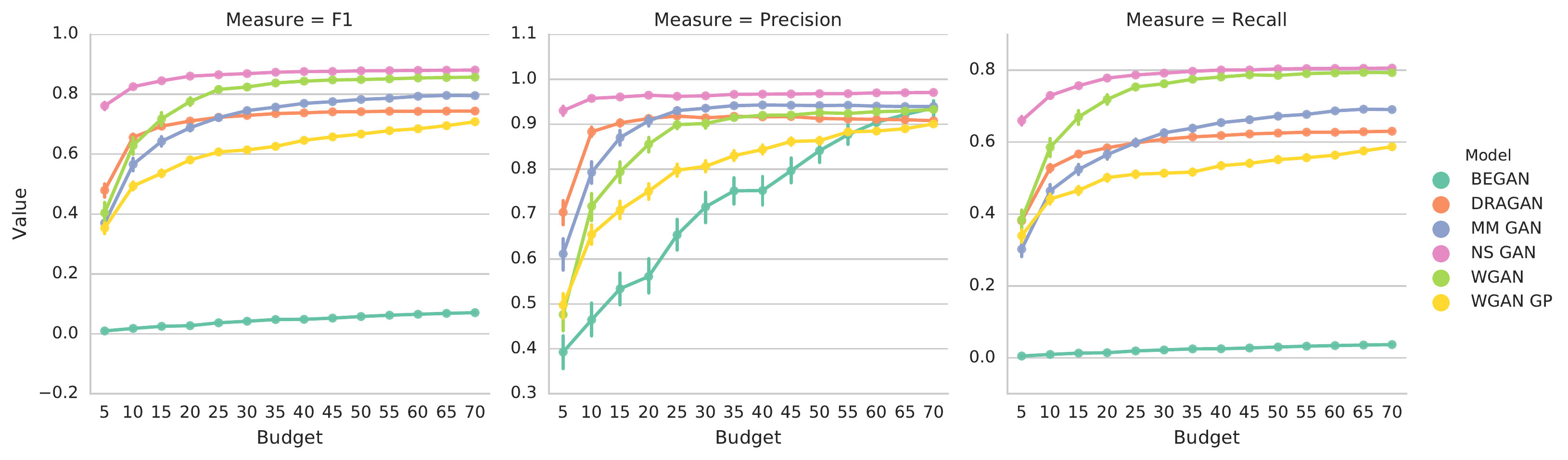}
  \caption{\small How does $F_1$ score vary with computational budget?
    The plot shows the distribution of the maximum $F_1$ score achievable
    for a fixed budget with a 95\% confidence interval. For each budget,
    we estimate the mean and confidence interval (of the mean) using
    $5000$ bootstrap resamples out of 100 runs. When optimizing for $F_1$
    score, both \textsc{ns gan} and \textsc{wgan} enjoy high precision and recall.
    The underwhelming performance of \textsc{began} and \textsc{vae} on this
    particular data set merits further investigation.\vspace{-5mm}}
  \label{fig:f1_vs_budget_wide}
\end{figure*}
We perform a search over the wide range of hyperparameters and compute
precision and recall by considering $n=1024$ samples.  In particular,
we compute the precision of the model by computing the fraction of
generated samples with distance below a threshold $\delta = 0.75$.
We then consider $n$ samples from the test set and invert each sample $x$ to compute $z^\star = G^{-1}(x)$ and compute the squared Euclidean distance
between $x$ and $G(z^\star)$. We define the recall as the
fraction of samples with squared Euclidean distance below $\delta$.
Figure~\ref{fig:f1_vs_budget_wide} shows the results where we select
the best $F_1$ score for a fixed model and hyperparameters and vary
the budget. We observe that even for this seemingly simple task, many
models struggle to achieve a high $F_1$ score. Analogous plots
where we instead maximize precision or recall for various thresholds
are presented in Appendix~\ref{app:f1_vs_budget}.

\section{Limitations of the Study}\label{sec:limitations}
\textbf{Data sets, neural architectures, and optimization issues.}
While we consider classic data sets from GAN research, unconditional
generation was recently applied to data sets of higher resolution and
arguably higher complexity. In this study we use one neural network
architecture which suffices to achieve good results in terms of FID on
all considered data sets.  However, given data sets of higher
complexity and higher resolution, it might be necessary to
significantly increase the number of parameters, which in turn might
lead to larger quantitative differences between different methods.
Furthermore, different objective functions might become sensible to
the choice of the optimization method, the number of training steps,
and possibly other optimization hyperparameters. These effects should
be systematically studied in future work.

\textbf{Metrics.}
It remains to be examined whether FID is stable under a more radical
change of the encoding, e.g using a network trained on a different
task. Furthermore, it might be possible to ``fool'' FID can probably by
introducing artifacts specialized to the encoding network. From the classic
machine learning point of view, a major drawback of FID
is that it cannot detect overfitting to the training data set --
an algorithm that outputs only the training examples would
have an excellent score. As such, developing quantitative
evaluation metrics is a critical research
direction~\cite{arora2018do,sajjadi2018assessing}.

\textbf{Exploring the space of hyperparameters.}
Ideally, hyperparameter values suggested by the authors should
transfer across data sets. As such, exploring the hyperparameters
"close" to the suggested ones is a natural and valid approach.
However, Figure~\ref{fig:phase1_boxplot} and in particular
Figure~\ref{fig:phase2_boxplot} show that optimization is
necessary. In addition, such an approach has several drawbacks: (a) no
recommended hyperparameters are available for a given data set, (b)
the parameters are different for each data set, (c) several popular models
have been tuned by the community, which might imply an unfair
comparison.  Finally, instead of random search it might be beneficial
to apply (carefully tuned) sequential Bayesian optimization which is
computationally beyond the scope of this study, but nevertheless a
great candidate for future work~\cite{kurach2018gan}.

\section{Conclusion}
In this paper we have started a discussion on how to neutrally and
fairly compare GANs. We focus on two sets of evaluation metrics: (i)
The Fr\'echet Inception Distance, and (ii) precision, recall and
$F_1$.  We provide empirical evidence that FID is a reasonable metric
due to its robustness with respect to mode dropping and encoding
network choices. Our main insight is that to compare models it is
meaningless to report the \emph{minimum} FID achieved.  Instead,
we propose to compare distributions of the minimum achivable FID for a fixed computational budget.
Indeed, empirical evidence presented herein imply that
algorithmic differences in state-of-the-art GANs become less relevant,
as the computational budget increases. Furthermore, given a limited
budget (say a month of compute-time), a ``good'' algorithm might be
outperformed by a ``bad'' algorithm.

As discussed in Section \ref{sec:comparison}, many dimensions have to
be taken into account for model comparison, and this work
only explores a subset of the options. We cannot exclude the
possibility that that some models significantly outperform others
under currently unexplored conditions. Nevertheless, notwithstanding the
limitations discussed in Section~\ref{sec:limitations}, this work
strongly suggests that future GAN research should be more
experimentally systematic and model comparison should be performed on neutral
ground.

\section*{Acknowledgments}
We would like to acknowledge Tomas Angles for advocating convex
polygons as a benchmark data set. We would like to thank Ian Goodfellow,
Michaela Rosca, Ishaan Gulrajani, David Berthelot, and Xiaohua Zhai for useful
discussions and remarks.

{
  \bibliographystyle{plainnat}
  \bibliography{paper}

\begin{thebibliography}{27}
\providecommand{\natexlab}[1]{#1}
\providecommand{\url}[1]{\texttt{#1}}
\expandafter\ifx\csname urlstyle\endcsname\relax
  \providecommand{\doi}[1]{doi: #1}\else
  \providecommand{\doi}{doi: \begingroup \urlstyle{rm}\Url}\fi

\bibitem[Arjovsky et~al.(2017)Arjovsky, Chintala, and
  Bottou]{arjovsky2017wasserstein}
Mart{\'{\i}}n Arjovsky, Soumith Chintala, and L{\'{e}}on Bottou.
\newblock Wasserstein generative adversarial networks.
\newblock In \emph{International Conference on Machine Learning (ICML)}, 2017.

\bibitem[Arora et~al.(2017)Arora, Ge, Liang, Ma, and
  Zhang]{arora2017generalization}
Sanjeev Arora, Rong Ge, Yingyu Liang, Tengyu Ma, and Yi~Zhang.
\newblock Generalization and equilibrium in generative adversarial nets
  ({GAN}s).
\newblock In \emph{International Conference on Machine Learning (ICML)}, 2017.

\bibitem[Arora et~al.(2018)Arora, Risteski, and Zhang]{arora2018do}
Sanjeev Arora, Andrej Risteski, and Yi~Zhang.
\newblock Do {GAN}s learn the distribution? some theory and empirics.
\newblock In \emph{International Conference on Learning Representations
  (ICLR)}, 2018.

\bibitem[Bachman and Precup(2015)]{bachman2015variational}
Philip Bachman and Doina Precup.
\newblock Variational generative stochastic networks with collaborative
  shaping.
\newblock In \emph{International Conference on Machine Learning (ICML)}, 2015.

\bibitem[Berthelot et~al.(2017)Berthelot, Schumm, and Metz]{berthelot2017began}
David Berthelot, Tom Schumm, and Luke Metz.
\newblock {BEGAN}: Boundary equilibrium generative adversarial networks.
\newblock \emph{arXiv preprint arXiv:1703.10717}, 2017.

\bibitem[Chen et~al.(2016)Chen, Chen, Duan, Houthooft, Schulman, Sutskever, and
  Abbeel]{chen2016infogan}
Xi~Chen, Xi~Chen, Yan Duan, Rein Houthooft, John Schulman, Ilya Sutskever, and
  Pieter Abbeel.
\newblock Infogan: Interpretable representation learning by information
  maximizing generative adversarial nets.
\newblock In \emph{Advances in {N}eural {I}nformation {P}rocessing {S}ystems
  (NIPS)}, 2016.

\bibitem[Fedus et~al.(2018)Fedus, Rosca, Lakshminarayanan, Dai, Mohamed, and
  Goodfellow]{fedus2017many}
William Fedus, Mihaela Rosca, Balaji Lakshminarayanan, Andrew~M. Dai, Shakir
  Mohamed, and Ian Goodfellow.
\newblock Many paths to equilibrium: {GAN}s do not need to decrease a
  divergence at every step.
\newblock In \emph{International Conference on Learning Representations
  (ICLR)}, 2018.

\bibitem[Gerhard et~al.(2013)Gerhard, Wichmann, and
  Bethge]{gerhard2013sensitive}
Holly~E Gerhard, Felix~A Wichmann, and Matthias Bethge.
\newblock How sensitive is the human visual system to the local statistics of
  natural images?
\newblock \emph{PLoS computational biology}, 9\penalty0 (1), 2013.

\bibitem[Goodfellow et~al.(2014)Goodfellow, Pouget-Abadie, Mirza, Xu,
  Warde-Farley, Ozair, Courville, and Bengio]{goodfellow2014generative}
Ian Goodfellow, Jean Pouget-Abadie, Mehdi Mirza, Bing Xu, David Warde-Farley,
  Sherjil Ozair, Aaron Courville, and Yoshua Bengio.
\newblock Generative adversarial nets.
\newblock In \emph{Advances in Neural Information Processing Systems (NIPS)},
  2014.

\bibitem[Gulrajani et~al.(2017)Gulrajani, Ahmed, Arjovsky, Dumoulin, and
  Courville]{gulrajani2017improved}
Ishaan Gulrajani, Faruk Ahmed, Martin Arjovsky, Vincent Dumoulin, and Aaron~C
  Courville.
\newblock Improved training of {W}asserstein gans.
\newblock In \emph{Advances in Neural Information Processing Systems (NIPS)},
  2017.

\bibitem[Heusel et~al.(2017)Heusel, Ramsauer, Unterthiner, Nessler, and
  Hochreiter]{heusel2017gans}
Martin Heusel, Hubert Ramsauer, Thomas Unterthiner, Bernhard Nessler, and Sepp
  Hochreiter.
\newblock {GAN}s trained by a two time-scale update rule converge to a local
  {N}ash equilibrium.
\newblock In \emph{Advances in Neural Information Processing Systems}, 2017.

\bibitem[Husz{\'a}r(2015)]{huszar2015not}
Ferenc Husz{\'a}r.
\newblock How (not) to train your generative model: Scheduled sampling,
  likelihood, adversary?
\newblock \emph{arXiv preprint arXiv:1511.05101}, 2015.

\bibitem[Kingma and Ba(2015)]{kingma2014adam}
Diederik~P Kingma and Jimmy Ba.
\newblock Adam: A method for stochastic optimization.
\newblock \emph{International Conference on Learning Representations (ICLR)},
  2015.

\bibitem[Kingma and Welling(2014)]{kingma2013auto}
Diederik~P Kingma and Max Welling.
\newblock Auto-encoding variational {B}ayes.
\newblock \emph{International Conference on Learning Representations (ICLR)},
  2014.

\bibitem[Kodali et~al.(2017)Kodali, Abernethy, Hays, and
  Kira]{kodali2017dragan}
Naveen Kodali, Jacob Abernethy, James Hays, and Zsolt Kira.
\newblock On convergence and stability of {GAN}s.
\newblock \emph{arXiv preprint arXiv:1705.07215}, 2017.

\bibitem[Kurach et~al.(2018)Kurach, Lucic, Zhai, Michalski, and
  Gelly]{kurach2018gan}
Karol Kurach, Mario Lucic, Xiaohua Zhai, Marcin Michalski, and Sylvain Gelly.
\newblock The {GAN} {L}andscape: {L}osses, architectures, regularization, and
  normalization.
\newblock \emph{arXiv preprint arXiv:1807.04720}, 2018.

\bibitem[Mahendran and Vedaldi(2015)]{mahendran2015understanding}
Aravindh Mahendran and Andrea Vedaldi.
\newblock Understanding deep image representations by inverting them.
\newblock In \emph{Conference on Computer Vision and Pattern Recognition
  (CVPR)}, 2015.

\bibitem[Mao et~al.(2017)Mao, Li, Xie, Lau, Wang, and Smolley]{mao2016least}
Xudong Mao, Qing Li, Haoran Xie, Raymond~YK Lau, Zhen Wang, and Stephen~Paul
  Smolley.
\newblock Least squares generative adversarial networks.
\newblock In \emph{International Conference on Computer Vision (ICCV)}, 2017.

\bibitem[Mescheder et~al.(2017)Mescheder, Nowozin, and
  Geiger]{mescheder2017numerics}
Lars Mescheder, Sebastian Nowozin, and Andreas Geiger.
\newblock The numerics of {GAN}s.
\newblock In \emph{Advances in Neural Information Processing Systems (NIPS)},
  2017.

\bibitem[Mirza and Osindero(2014)]{mirza2014conditional}
Mehdi Mirza and Simon Osindero.
\newblock Conditional generative adversarial nets.
\newblock \emph{arXiv preprint arXiv:1411.1784}, 2014.

\bibitem[Odena et~al.(2017)Odena, Olah, and Shlens]{odena2016conditional}
Augustus Odena, Christopher Olah, and Jonathon Shlens.
\newblock Conditional image synthesis with auxiliary classifier {GAN}s.
\newblock In \emph{International Conference on Machine Learning (ICML)}, 2017.

\bibitem[Radford et~al.(2015)Radford, Metz, and
  Chintala]{radford2015unsupervised}
Alec Radford, Luke Metz, and Soumith Chintala.
\newblock Unsupervised representation learning with deep convolutional
  generative adversarial networks.
\newblock \emph{arXiv preprint arXiv:1511.06434}, 2015.

\bibitem[Sajjadi et~al.(2018)Sajjadi, Bachem, Lucic, Bousquet, and
  Gelly]{sajjadi2018assessing}
Mehdi~SM Sajjadi, Olivier Bachem, Mario Lucic, Olivier Bousquet, and Sylvain
  Gelly.
\newblock Assessing generative models via precision and recall.
\newblock In \emph{Advances in Neural Information Processing Systems (NIPS)},
  2018.

\bibitem[Salimans et~al.(2016)Salimans, Goodfellow, Zaremba, Cheung, Radford,
  and Chen]{salimans2016improved}
Tim Salimans, Ian Goodfellow, Wojciech Zaremba, Vicki Cheung, Alec Radford, and
  Xi~Chen.
\newblock Improved techniques for training {GAN}s.
\newblock In \emph{Advances in {N}eural {I}nformation {P}rocessing {S}ystems
  (NIPS)}, 2016.

\bibitem[Theis et~al.(2015)Theis, Oord, and Bethge]{theis2015note}
Lucas Theis, A{\"a}ron van~den Oord, and Matthias Bethge.
\newblock A note on the evaluation of generative models.
\newblock \emph{arXiv preprint arXiv:1511.01844}, 2015.

\bibitem[Wu et~al.(2017)Wu, Burda, Salakhutdinov, and
  Grosse]{wu2016quantitative}
Yuhuai Wu, Yuri Burda, Ruslan Salakhutdinov, and Roger Grosse.
\newblock On the quantitative analysis of decoder-based generative models.
\newblock \emph{International Conference on Learning Representations (ICLR)},
  2017.

\bibitem[Zhang et~al.(2017)Zhang, Xu, Li, Zhang, Huang, Wang, and
  Metaxas]{zhang2016stackgan}
Han Zhang, Tao Xu, Hongsheng Li, Shaoting Zhang, Xiaolei Huang, Xiaogang Wang,
  and Dimitris Metaxas.
\newblock Stackgan: Text to photo-realistic image synthesis with stacked
  generative adversarial networks.
\newblock \emph{International Conference on Computer Vision (ICCV)}, 2017.

\end{thebibliography}
}

\onecolumn
\appendix
\section{Wide and narrow hyperparameter ranges}\label{app:hyperparameters}

The \emph{wide} and \emph{narrow} ranges of hyper-parameters are presented in
Table~\ref{tab:parameter_ranges_wide} and
Table~\ref{tab:parameter_ranges_narrow} respectively. In both tables, U(a, b)
means that the variable was sample uniformly from the range $[a, b]$.
The L(a, b) means that that the variable was sampled on a log-scale,
that is $x ~ L(a, b) \iff x ~ 10^{U(log(a), log(b))}$.
The parameters used in the search:
\begin{itemize}
\item $\beta_1$: the parameter of the Adam optimization algorithm.
\item Learning rate: generator/discriminator learning rate.
\item $\lambda$: Multiplier of the gradient penalty for \textsc{dragan} and \textsc{wgan gp}. Learning rate for $k_t$ in  \textsc{began}.
\item Disc iters: Number of discriminator updates per one generator update.
\item batchnorm: If True, the batch normalization will be used in the discriminator.
\item $\gamma$: Parameter of \textsc{began}.
\item clipping: Parameter of \textsc{wgan}, weights will be clipped to this
  value.
\end{itemize}

\begin{table*}[h]
  \renewcommand{\arraystretch}{1.6}
  \centering
  \caption{\small Wide ranges of hyper-parameters used for the large-scale search.
    ``U'' denotes uniform sampling, ``L'' sampling on a log-scale.
    \label{tab:parameter_ranges_wide}}
  {
    \scriptsize
    \begin{tabular}{lcccccccc}
\toprule
 & \textsc{mm gan} &  \textsc{ns gan} &  \textsc{lsgan} &  \textsc{wgan}
  & \textsc{wgan gp} & \textsc{dragan} & \textsc{began} & \textsc{vae}\\
\midrule
  Adam's $\beta_1$ & \multicolumn{8}{c}{ U($0, 1$) } \\
  learning rate &  \multicolumn{8}{c}{ L($10^{-5}, 10^{-2}$)} \\
  $\lambda$ & N/A & N/A & N/A & N/A & L($10^{-1}, 10^2$) & L($10^{-1}, 10^2$) & L($10^{-4}, 10^{-2}$) & N/A \\
  disc iter  &  \multicolumn{8}{c}{Either $1$ or $5$, sampled with the same probablity} \\
  batchnorm & \multicolumn{8}{c}{True or False, sampled with the same probability} \\
  $\gamma$ & N/A & N/A & N/A & N/A & N/A & N/A & U($0, 1$) & N/A \\
  clipping & N/A & N/A & N/A & L($10^{-3}, 10^0$) & N/A & N/A & N/A & N/A \\
\bottomrule
\end{tabular}

  }
\end{table*}

\begin{table*}[h]
  \small
  \renewcommand{\arraystretch}{1.4}
  \centering
  \caption{\small Narrow ranges of hyper-parameters used in the search with $50$
    samples per model. The ranges were optimized by looking at the wide search
    results for fashion-mnist data set.  ``U'' denotes uniform sampling, ``L''
    sampling on a log-scale.
    \label{tab:parameter_ranges_narrow}}
  {
    \scriptsize
    \begin{tabular}{lcccccccc}
\toprule
 & \textsc{mm gan} &  \textsc{ns gan} &  \textsc{lsgan} &  \textsc{wgan}
  & \textsc{wgan gp} & \textsc{dragan} & \textsc{began} & \textsc{vae}\\
\midrule
  Adam's $\beta_1$ & \multicolumn{8}{c}{ Always $0.5$} \\
  learning rate &  \multicolumn{8}{c}{ L($10^{-4}, 10^{-3}$)} \\
  $\lambda$ & N/A & N/A & N/A & N/A & L($10^{-1}, 10^1$) & L($10^{-1}, 10^1$) & L($10^{-4}, 10^{-2}$) & N/A \\
  disc iter  &  \multicolumn{8}{c}{Always $1$} \\
  batchnorm & True/False & True/False & True/False & True/False & False & False & True/False & True/False \\
  $\gamma$ & N/A & N/A & N/A & N/A & N/A & N/A & U($0.6, 0.9$) & N/A \\
  clipping & N/A & N/A & N/A & L($10^{-2}, 10^0$) & N/A & N/A & N/A & N/A \\
\bottomrule
\end{tabular}

  }
\end{table*}
\section{Which parameters really matter?}

\begin{figure*}[h]
  \begin{centering}
    \includegraphics[width=\columnwidth]{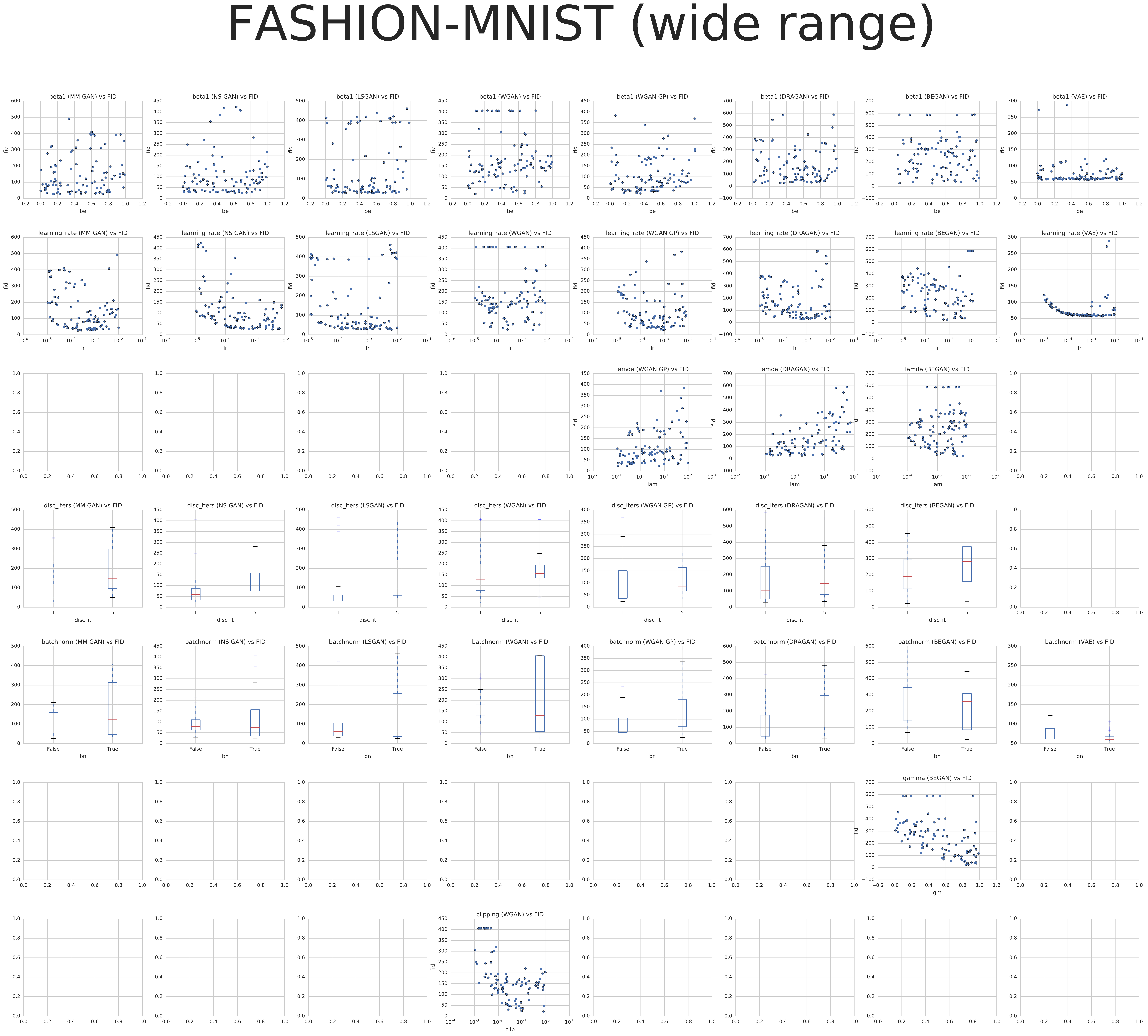}
    \caption{\small Wide range scatter plots for \textsc{fashion mnist}.
      For each algorithm (column) and each parameter (row), the corresponding scatter plot shows the FID
      in function of the parameter. This illustrates the sensitivity of each algorithm
      w.r.t. each parameter. Those results have been used to choose the \emph{narrow range}
      in Table~\ref{tab:parameter_ranges_narrow}. For example, Adam's $\beta_1$ does not seem to
      significantly impact any algorithm, so for the narrow range, we fix its value to always be $0.5$.
      Likewise, we fix the number of discriminator iterations to be always 1. For other parameters, the selected range
      is smaller (e.g. learning rate), or can differ for each algorithm (e.g. batch norm). \label{fig:phase1_fashion_scatter}}
  \end{centering}
\end{figure*}

\begin{figure*}[h]
  \centering
  \includegraphics[width=\columnwidth]{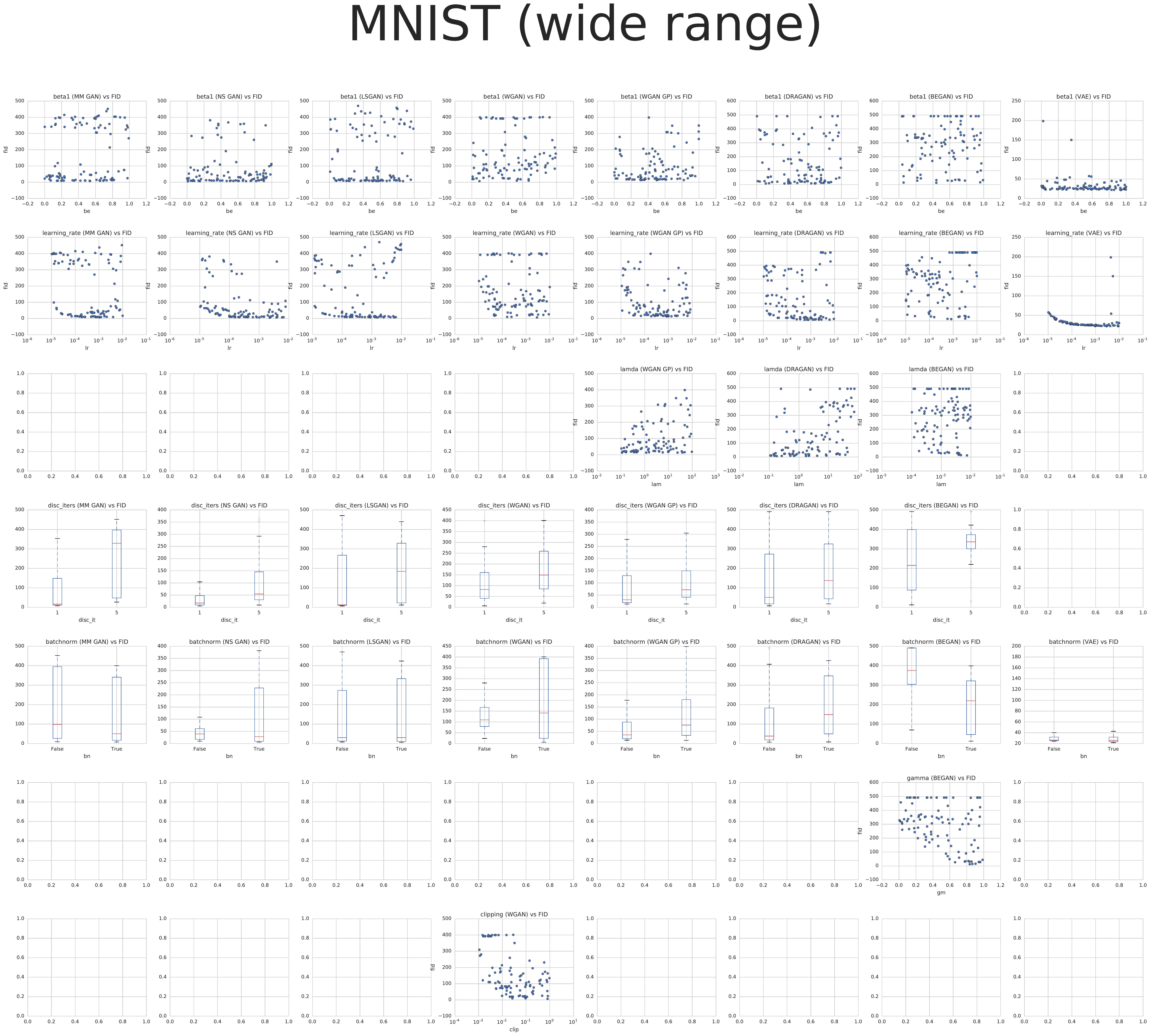}
  \caption{\small Wide range scatter plots for \textsc{mnist}.
    For each algorithm (column) and each parameter (row), the corresponding scatter plot shows the FID
    in function of the parameter. This illustrates the sensitivity of each algorithm
    w.r.t. each parameter.   \label{fig:phase1_mnist_scatter}
  }
\end{figure*}

\begin{figure*}[h]
  \centering
  \includegraphics[width=\columnwidth]{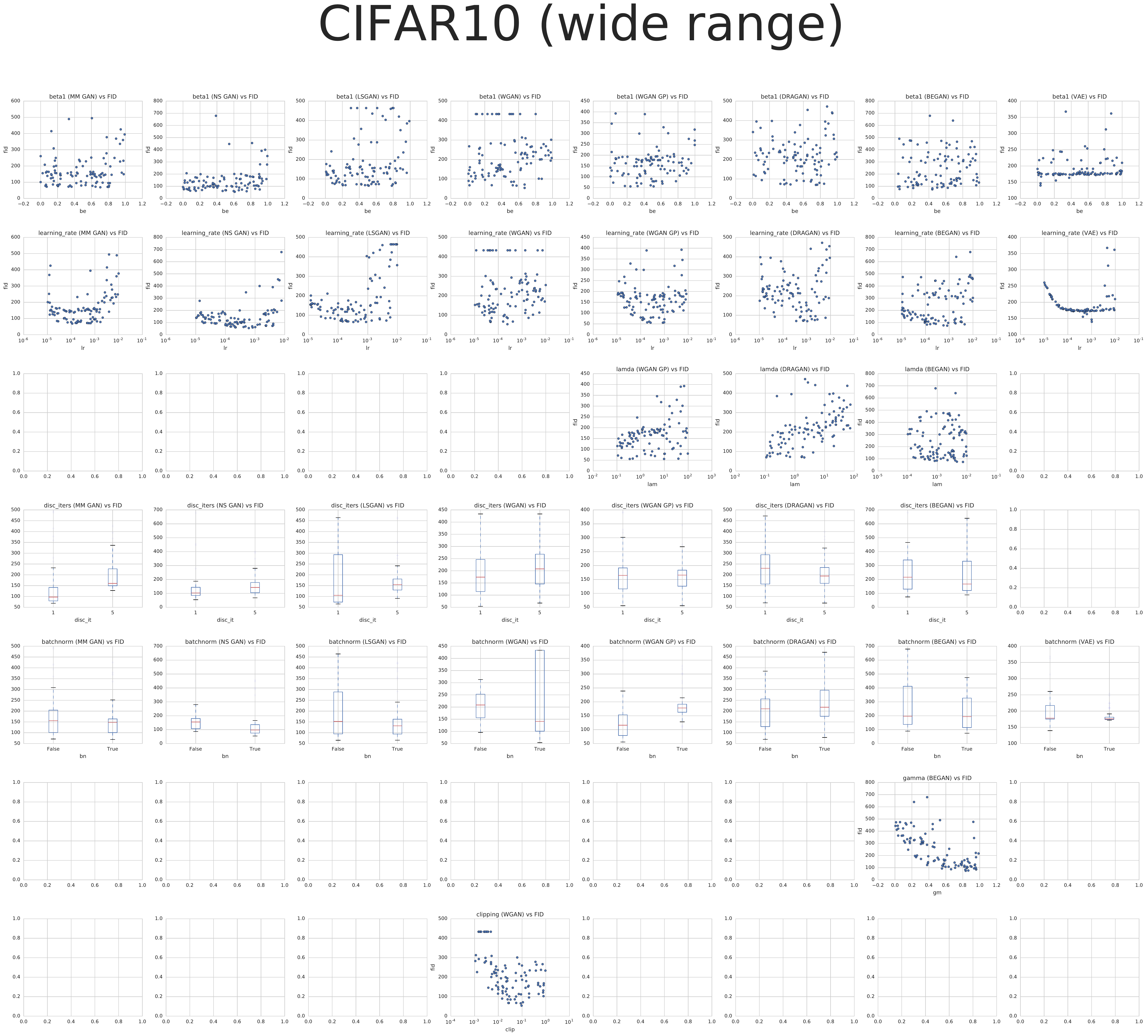}
  \caption{\small Wide range scatter plots for \textsc{cifar10}.
  For each algorithm (column) and each parameter (row), the corresponding scatter plot shows the FID
  in function of the parameter. This illustrates the sensitivity of each algorithm
  w.r.t. each parameter.
}
  \label{fig:phase1_cifar10_scatter}
\end{figure*}

\begin{figure*}[h]
  \centering
  \includegraphics[width=\columnwidth]{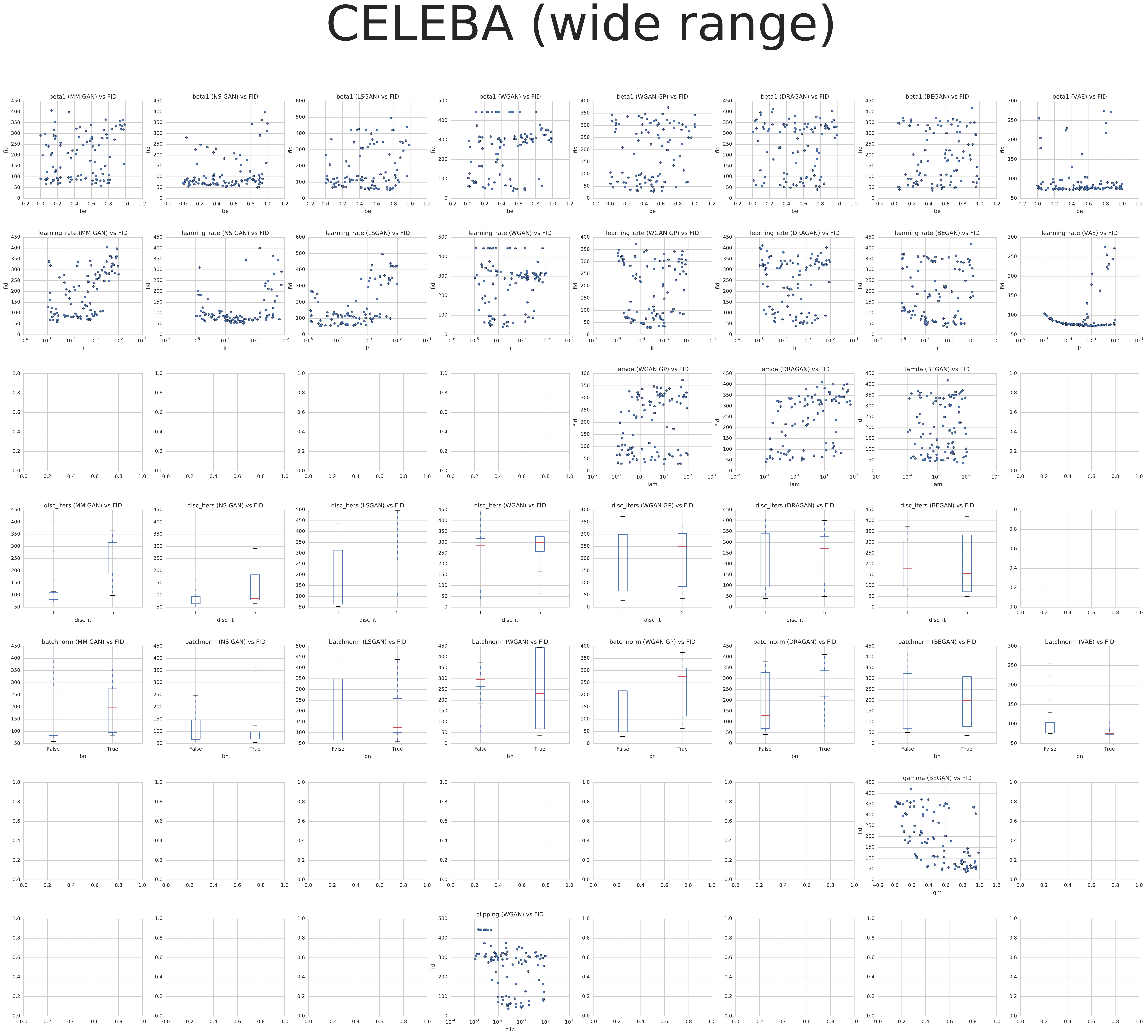}
  \caption{\small Wide range scatter plots for \textsc{CelebA}.
  For each algorithm (column) and each parameter (row), the corresponding scatter plot shows the FID
  in function of the parameter. This illustrates the sensitivity of each algorithm
  w.r.t. each parameter.
}

  \label{fig:phase1_celeba_scatter}
\end{figure*}

Figure~\ref{fig:phase1_fashion_scatter}, Figure~\ref{fig:phase1_mnist_scatter},
Figure~\ref{fig:phase1_cifar10_scatter} and Figure~\ref{fig:phase1_celeba_scatter}
present scatter plots for data sets \textsc{fashion mnist}, \textsc{mnist},
\textsc{cifar}, \textsc{CelebA} respectively. For each model and
hyper-parameter we estimate its impact on the final FID. Figure~\ref{fig:phase1_fashion_scatter} was used to select narrow ranges
of hyper-parameters.

\newpage

\newpage

\newpage
\newpage

\clearpage
\newpage

\newpage
\section{Fr\'echet Inception Distance and Image Quality}\label{app:fidwithimages}
\begin{figure*}[h]
  \centering
  \includegraphics[width=\columnwidth]{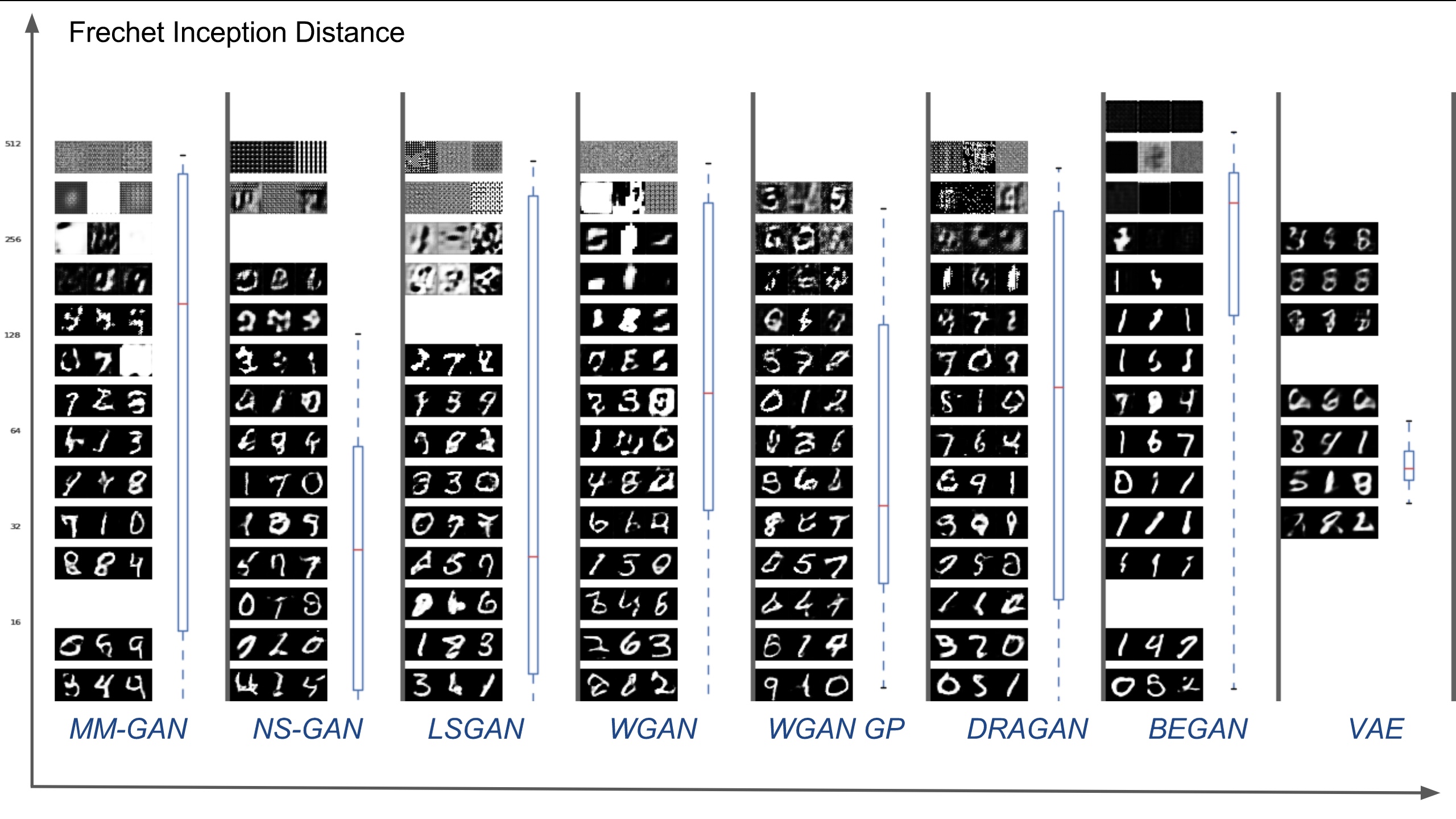}
  \caption{\small \textsc{mnist}: Distribution of FIDs and corresponding samples for each model when sampling parameters from \emph{wide} ranges.}
  \label{fig:fid_with_pictures_mnist_wide}
\end{figure*}

\begin{figure*}[h]
  \centering
  \includegraphics[width=\columnwidth]{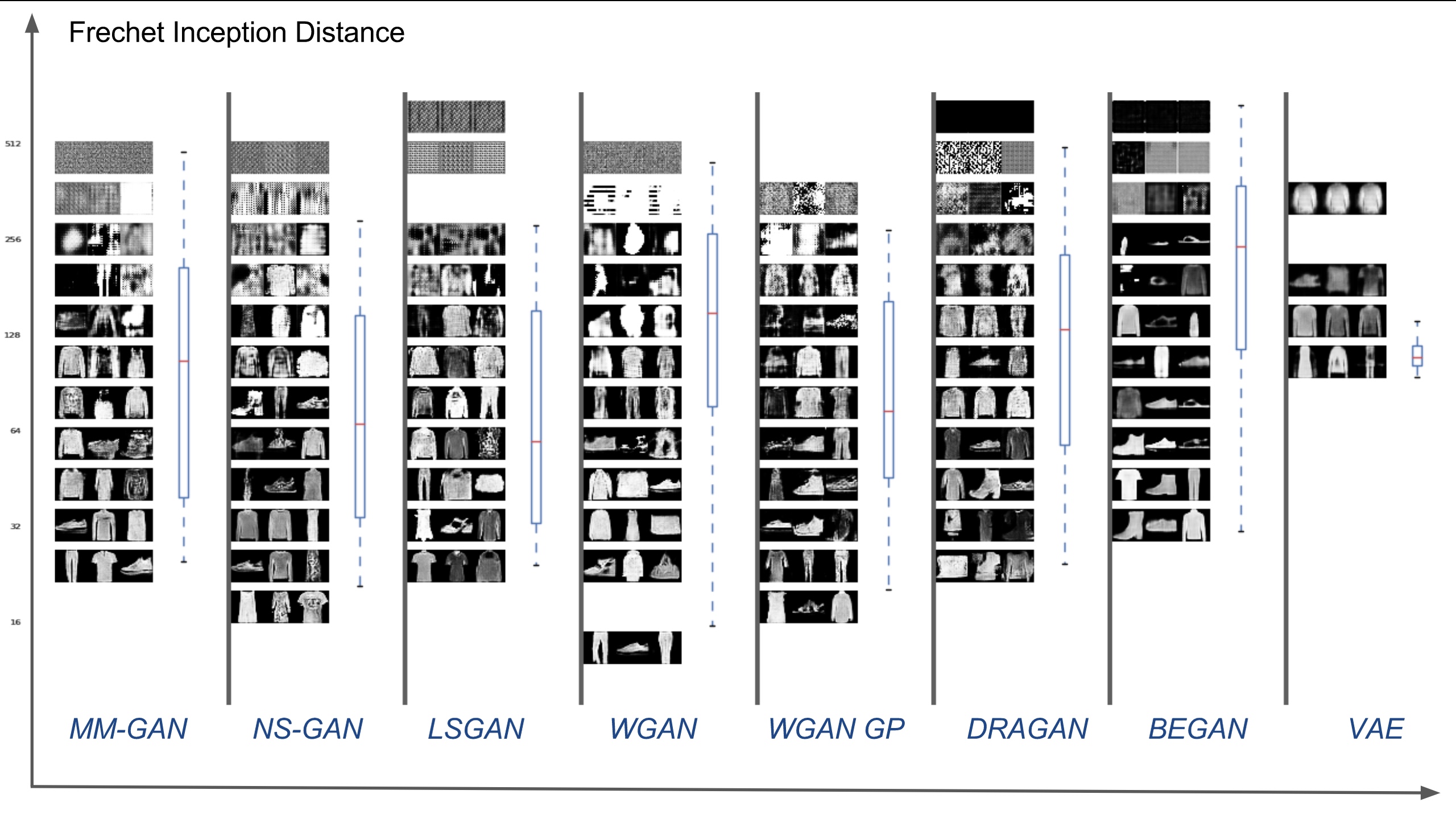}
  \caption{\small \textsc{fashion-mnist}: Distribution of FIDs and corresponding samples for each model when sampling parameters from \emph{wide} ranges.}
  \label{fig:fid_with_pictures_fashion_wide}
\end{figure*}

\begin{figure*}[h]
  \centering
  \includegraphics[width=\columnwidth]{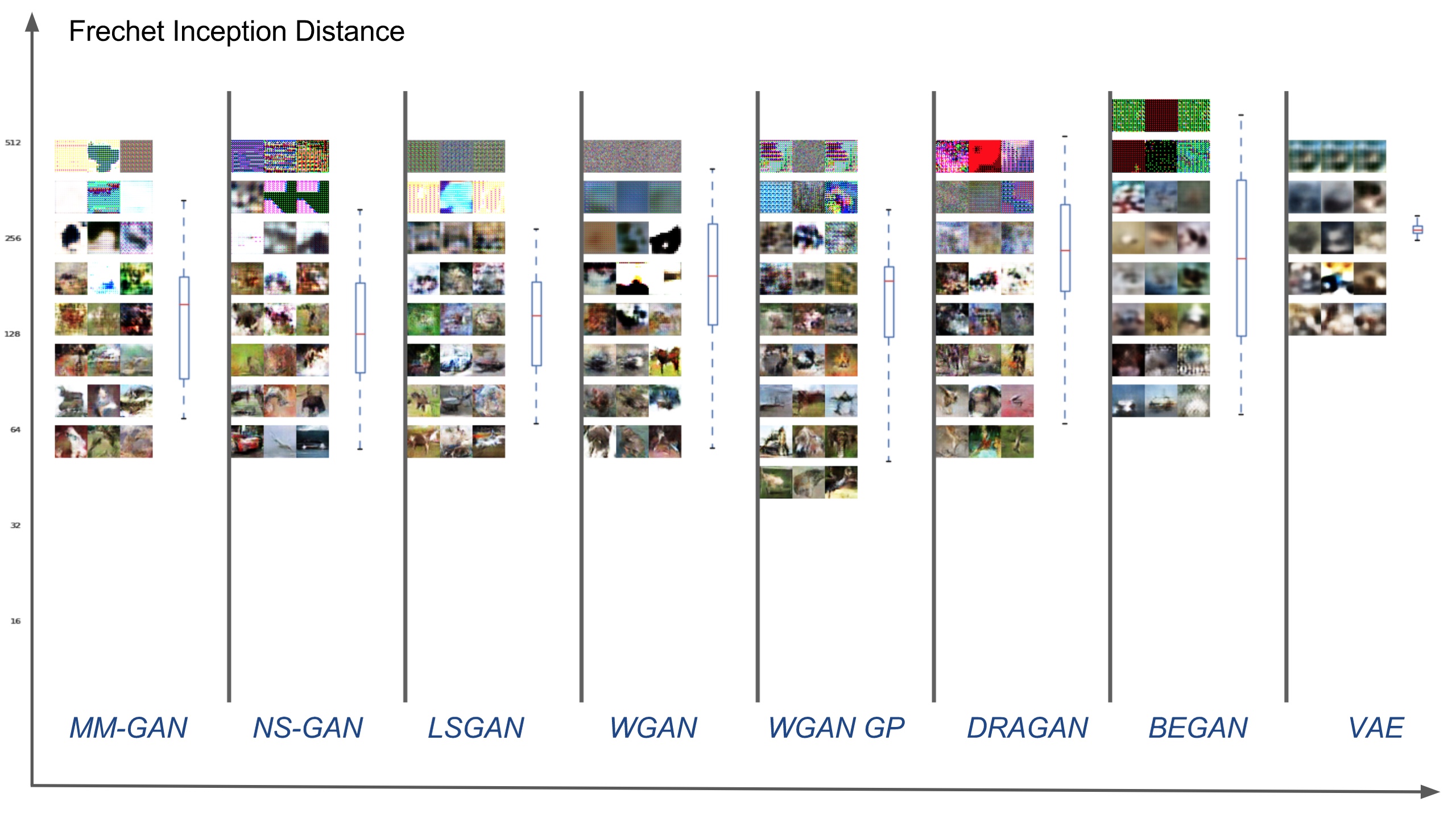}
  \caption{\small \textsc{cifar10}: Distribution of FIDs and corresponding samples for each model when sampling parameters from \emph{wide} ranges.}
  \label{fig:fid_with_pictures_cifar10_wide}
\end{figure*}

\begin{figure*}[h]
  \centering
  \includegraphics[width=\columnwidth]{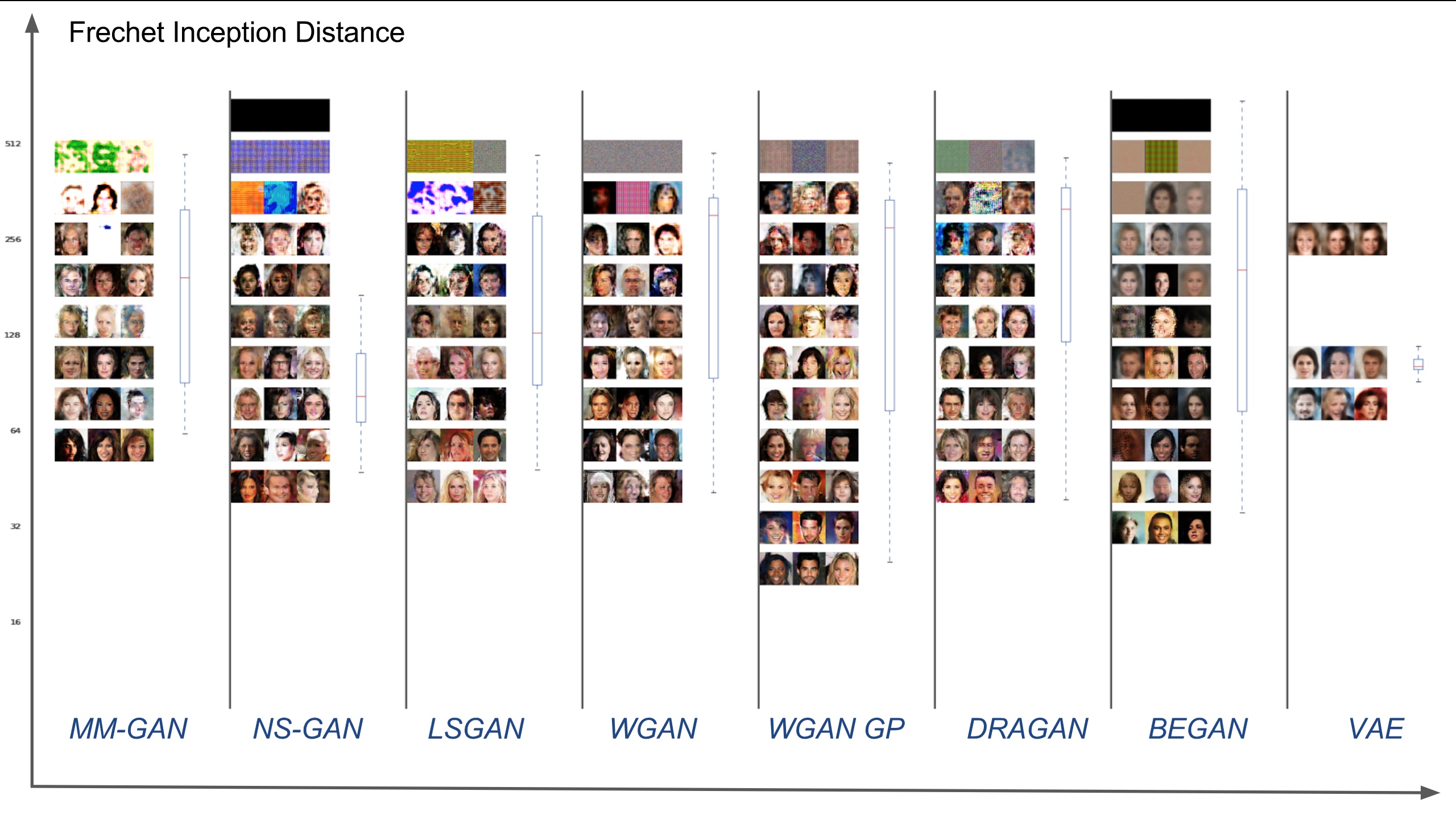}
  \caption{\small \textsc{celeba}: Distribution of FIDs and corresponding samples for each model when sampling parameters from \emph{wide} ranges.}
  \label{fig:fid_with_pictures_celeba_wide}
\end{figure*}

It is interesting to see how the FID translates to the image quality.
In Figure~\ref{fig:fid_with_pictures_mnist_wide},
Figure~\ref{fig:fid_with_pictures_fashion_wide},
Figure~\ref{fig:fid_with_pictures_cifar10_wide} and
Figure~\ref{fig:fid_with_pictures_celeba_wide}, we present, for every model,
the distribution of FIDs and the corresponding samples.


\newpage
\clearpage
\section{Hyper-parameter Search over Narrow Ranges}
\begin{figure*}[h]
  \centering
  \includegraphics[width=\columnwidth]{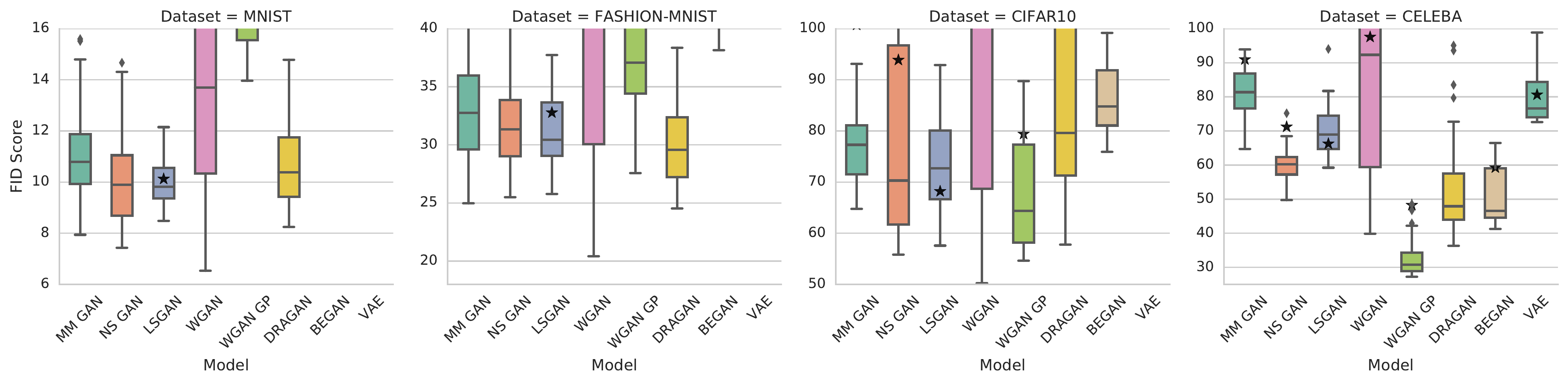}
  \caption{\small A \emph{narrow range} search of hyperparameters which were selected based on the wide hyperparameter search on
  the \textsc{fashion-mnist} data set. Black stars indicate the performance of suggested hyperparameter settings. For each model we allow $50$ hyperparameter samples. From the practical point of view, there are significant differences between the models: in some cases the hyperparameter ranges \emph{transfer} from one data set to the others (e.g.
  \textsc{ns gan}), while others are more sensitive to this choice (e.g.
  \textsc{wgan}).  We note that better scores can be obtained by a wider
  hyperparameter search. }
  \label{fig:phase2_boxplot}
\end{figure*}

In Figure~\ref{fig:phase1_boxplot} we presented the sensitivity of GANs to
hyperparameters, assuming the samples are taken from the wide ranges (see
Table~\ref{tab:parameter_ranges_wide}). For completeness, in
Figure~\ref{fig:phase2_boxplot} we present a similar comparison for the narrow
ranges of hyperparameters (presented in Table~\ref{tab:parameter_ranges_narrow}).

\begin{figure*}[h]
  \centering
  \includegraphics[width=\columnwidth]{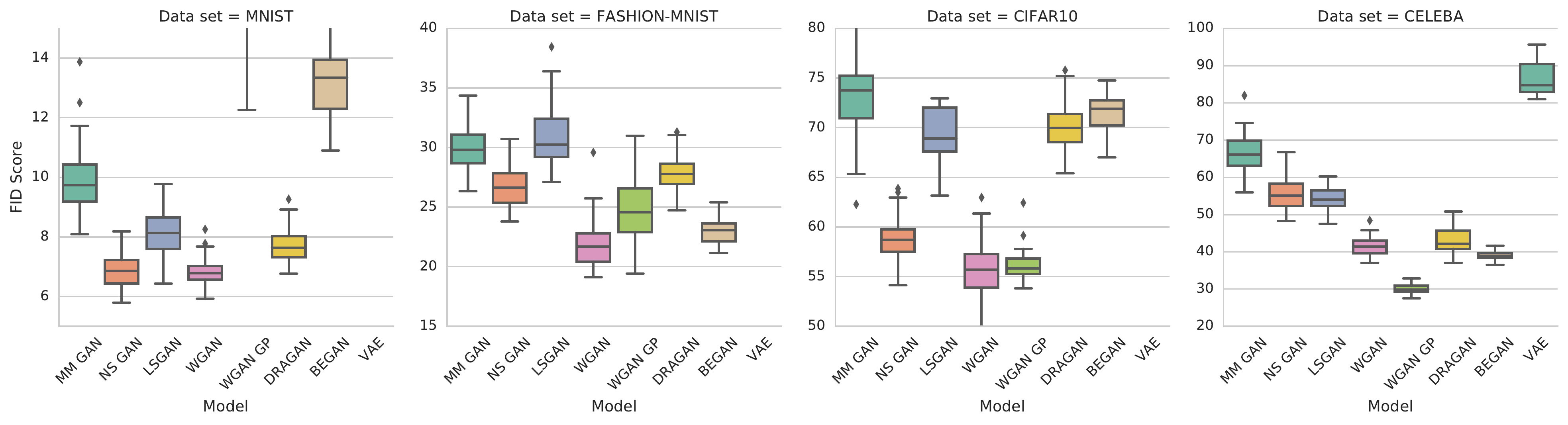}
  \caption{\small For each model we search for best hyperparameters on the
  \emph{wide range}. Then, we retrain each model using the best
  parameters $50$ times with random initialization of the weights,
  keeping everything else fixed. We observe a slight variance in the
  final FID. Hence, when an FID is reported it is paramount
  that one compares the entire distribution, instead of the \emph{best}
  seed for the best run. The figure corresponds to Table~\ref{tab:best_fid_per_dataset}.}
  \label{fig:fid_variance_fixed_hyperparameters}
\end{figure*}

\clearpage
\newpage
\section{Precision, Recall and $F_1$ as a Function of the Budget}\label{app:f1_vs_budget}
\begin{figure*}[h]
  \centering
  \includegraphics[width=\columnwidth]{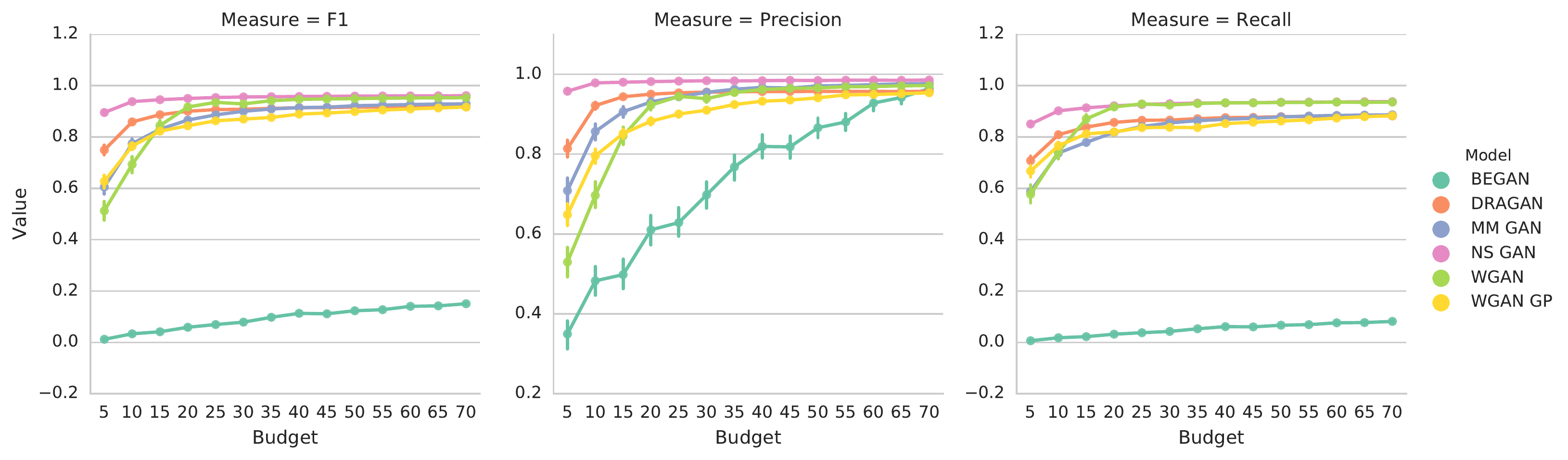}
  \caption{\small Optimizing for $F_1$, threshold $\delta=1.0$.}
  \label{fig:f1_vs_budget_wide_f1_3}
\end{figure*}
\begin{figure*}[h]
  \centering
  \includegraphics[width=\columnwidth]{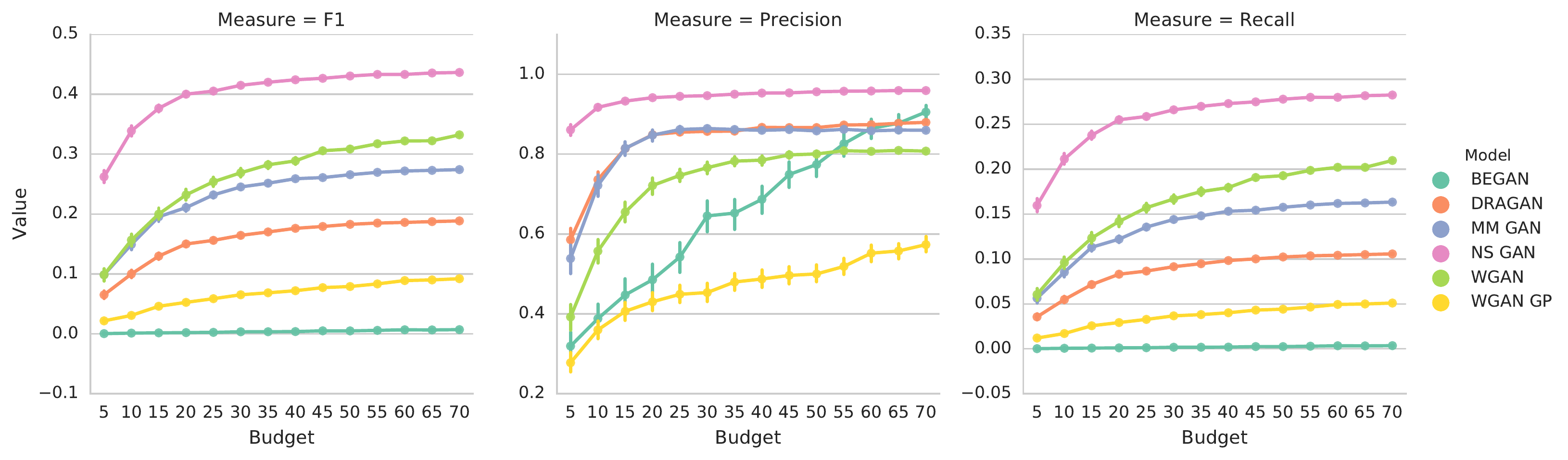}
  \caption{\small Optimizing for $F_1$, threshold $\delta=0.5$.}
  \label{fig:f1_vs_budget_wide_f1_5}
\end{figure*}

\begin{figure*}[h]
  \centering
  \includegraphics[width=\columnwidth]{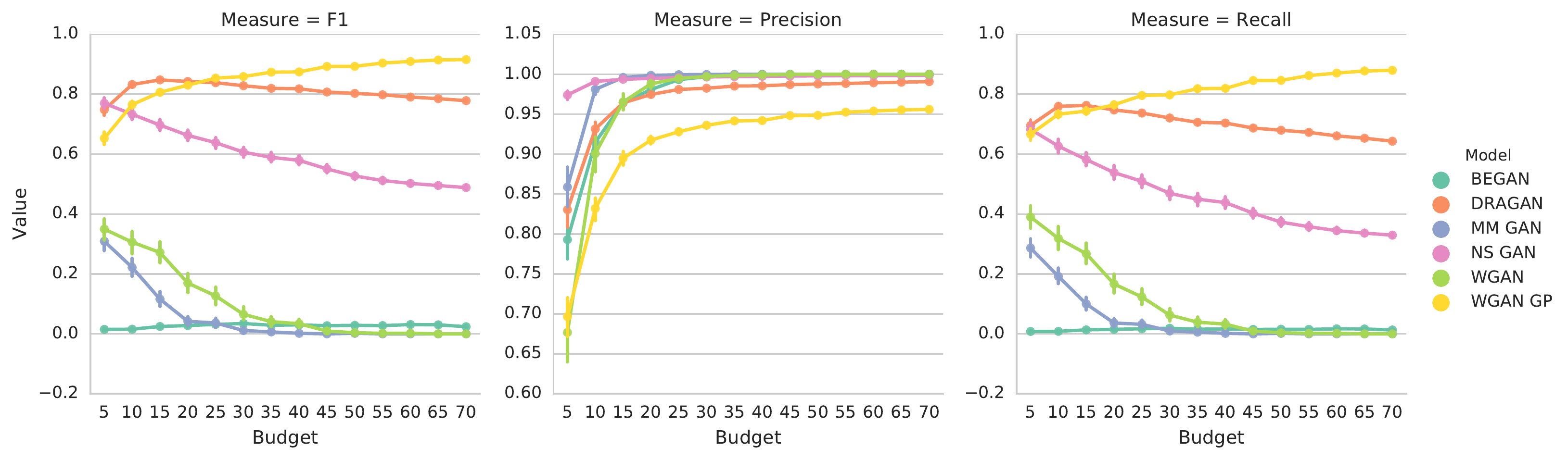}
  \caption{\small Optimizing for precision, threshold $\delta=1.0$.}
  \label{fig:f1_vs_budget_wide_precision_3}
\end{figure*}
\begin{figure*}[h]
  \centering
  \includegraphics[width=\columnwidth]{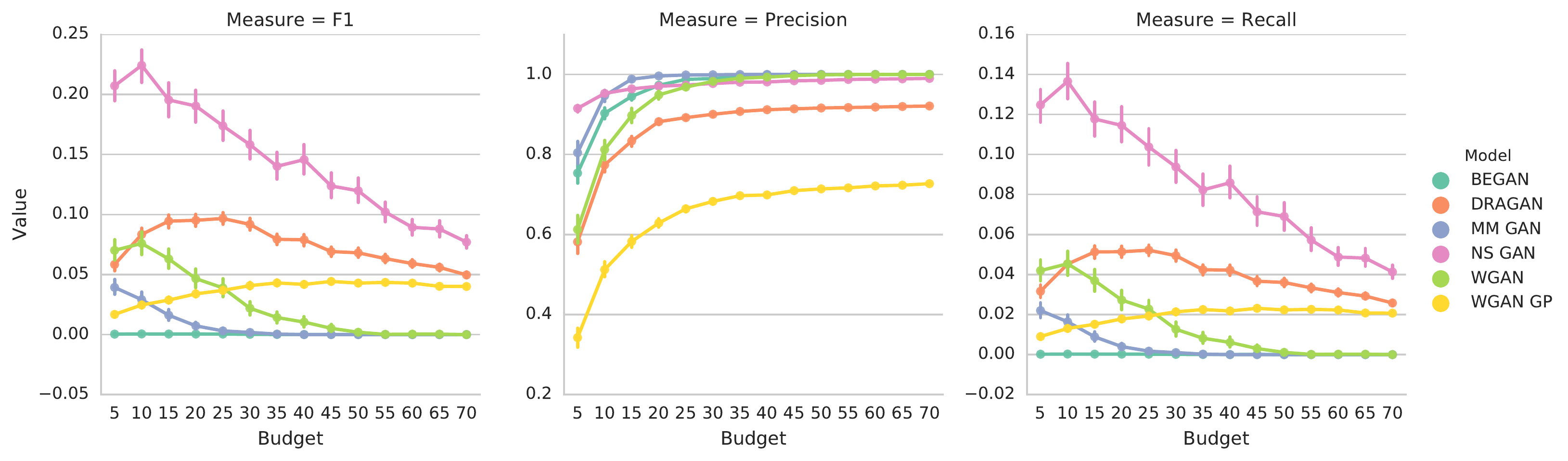}
  \caption{\small Optimizing for precision, threshold $\delta=0.5$.}
  \label{fig:f1_vs_budget_wide_precision_5}
\end{figure*}

\begin{figure*}[h]
  \centering
  \includegraphics[width=\columnwidth]{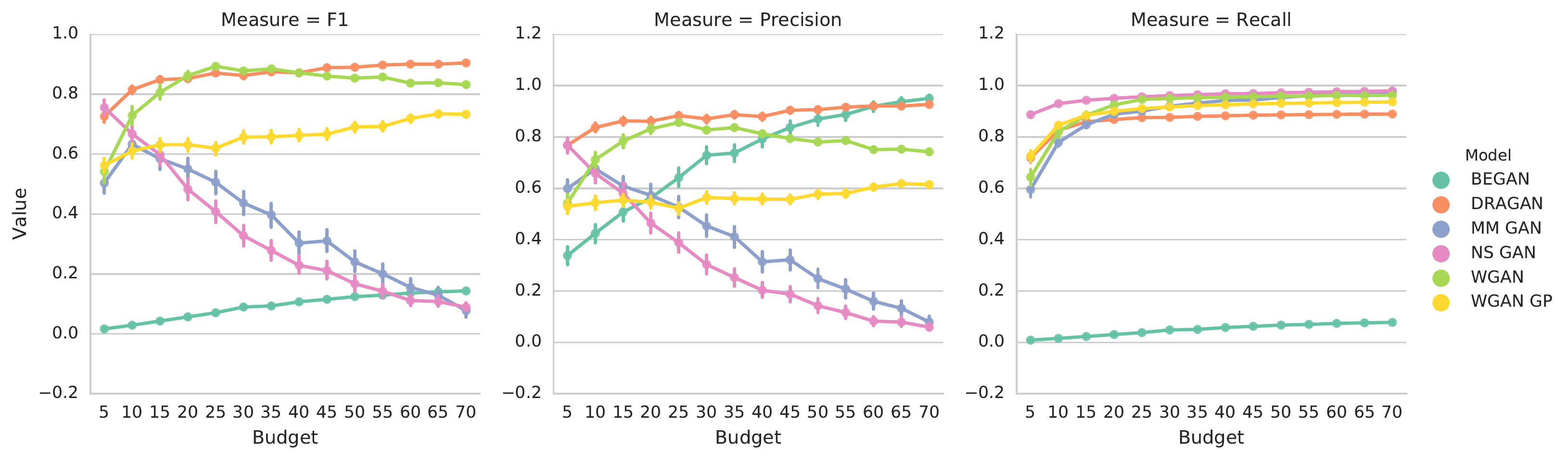}
  \caption{\small Optimizing for recall, threshold $\delta=1.0$.}
  \label{fig:f1_vs_budget_wide_recall_3}
\end{figure*}
\begin{figure*}[h]
  \centering
  \includegraphics[width=\columnwidth]{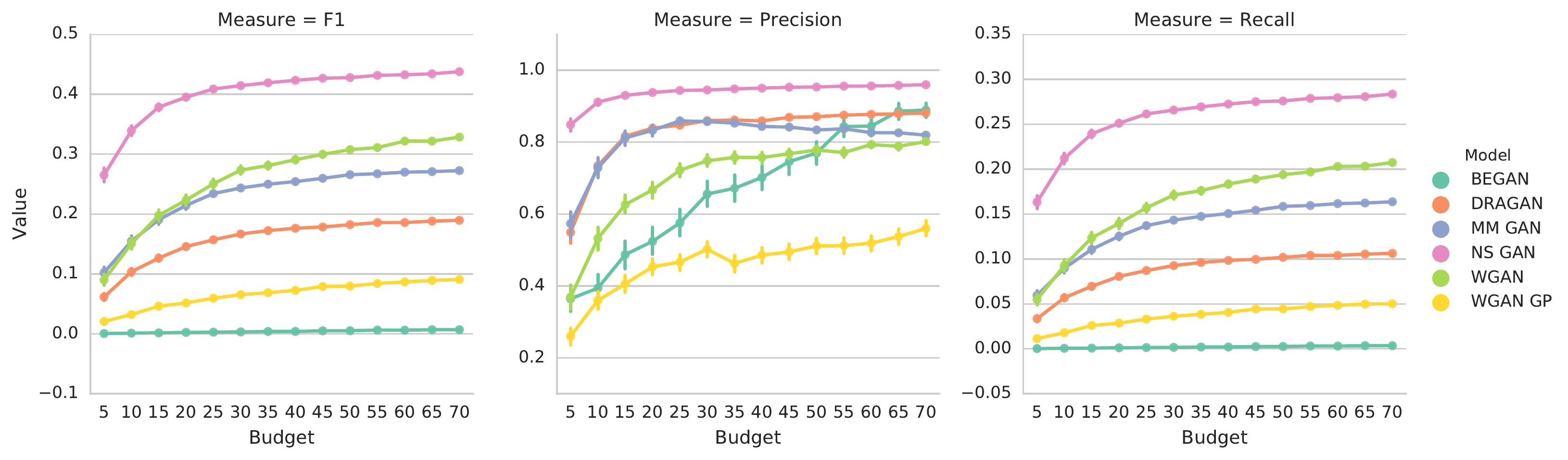}
  \caption{\small Optimizing for recall, threshold $\delta=0.5$.}
  \label{fig:f1_vs_budget_wide_recall_50-1024}
\end{figure*}
\section{Impact of the optimization algorithm.}\label{app:solver}
We ran the WGAN training across 100 hyperparameter settings. In the first set of experiments we used the \textsc{ADAM} optimizer, and in the second the \textsc{RMSProp} optimizer. We observe that distribution of the scores is similar and it's unclear which optimizer is "better". However, on both data sets \textsc{ADAM} outperformed \textsc{RMSProp} on recommended parameters (\textsc{cifar10:} 154.5 vs 161.2, \textsc{Celeba:} 97.9 vs 216.3) which highlights the need for a hyperparameter search. As a result, the conclusions of this work are not altered by this choice.

\clearpage
\section{$F_1$, precision, and recall correlation with FID}\label{app:correlation_fid_f1}
\begin{figure*}[h]
  \centering
  \includegraphics[width=.32\columnwidth]{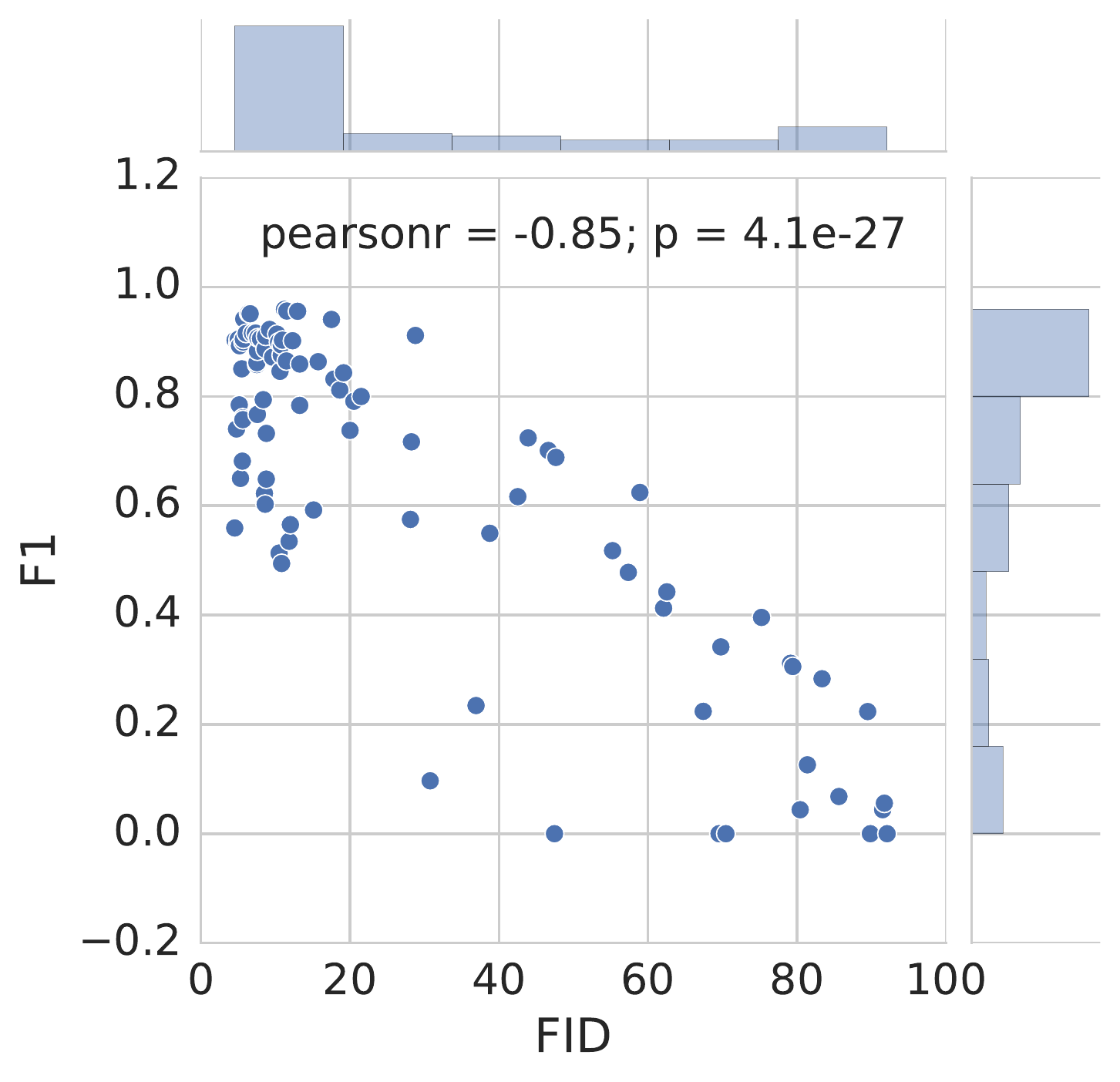}
  \includegraphics[width=.32\columnwidth]{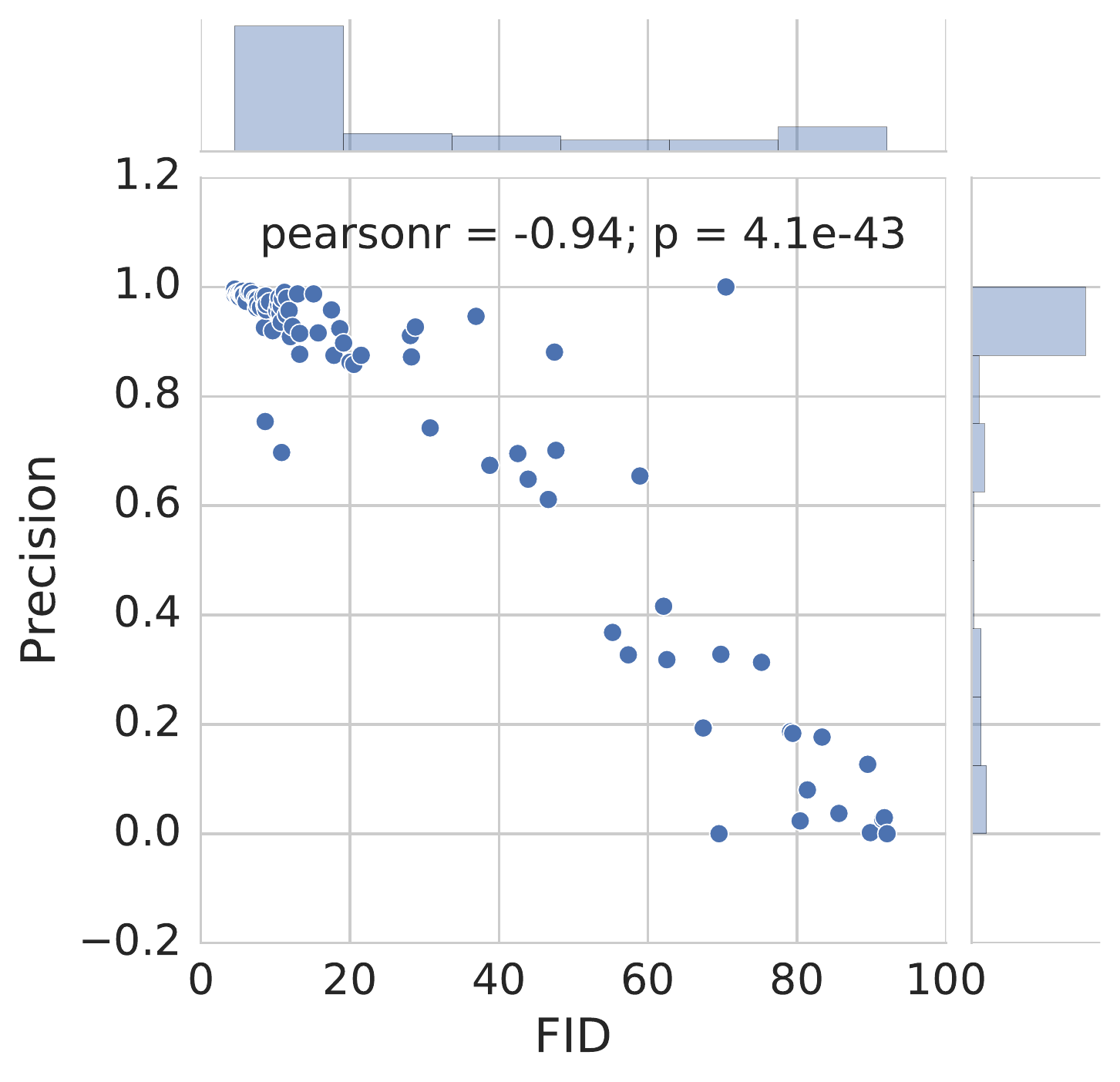}
  \includegraphics[width=.32\columnwidth]{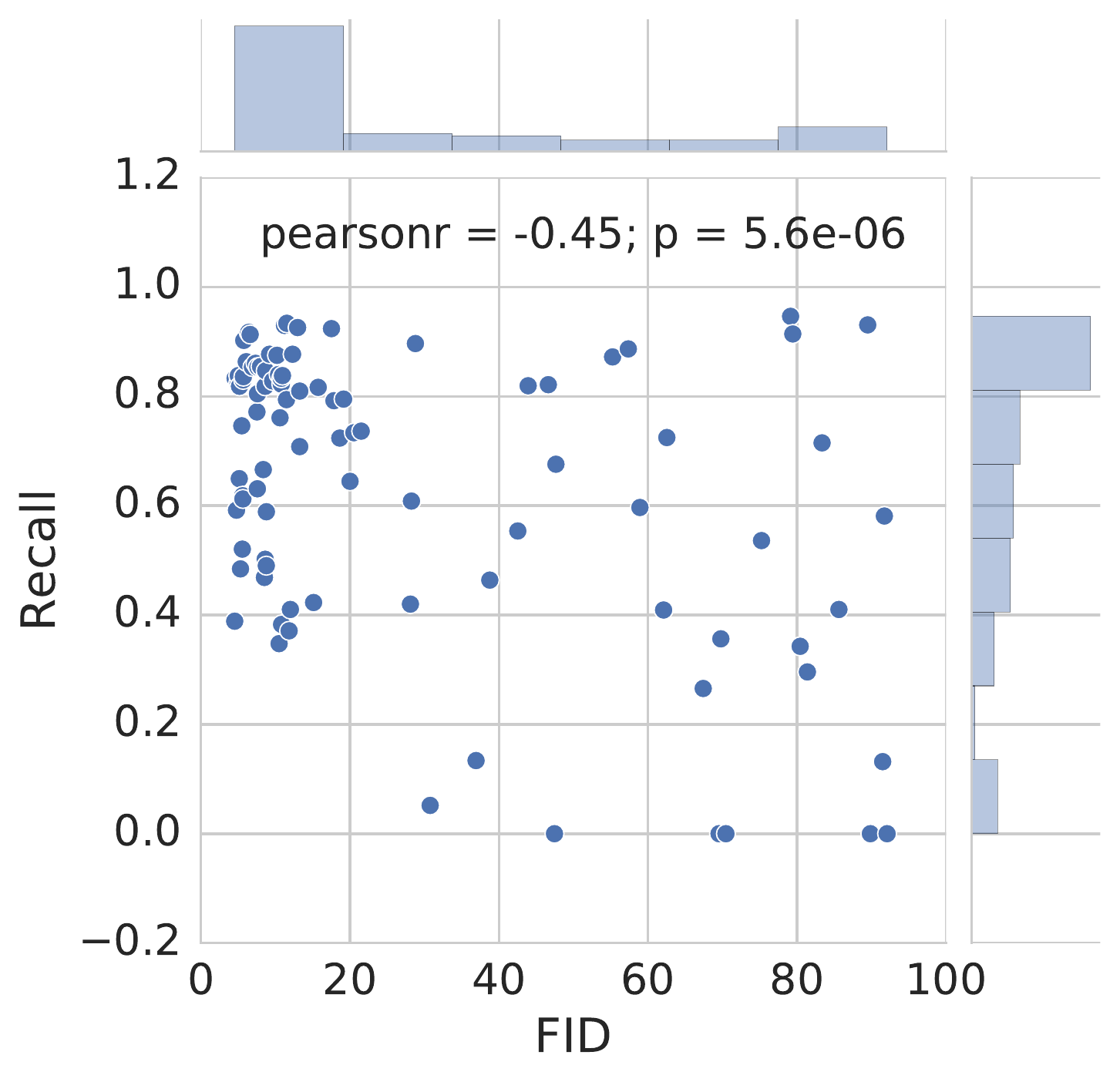}
  \caption{\small Correlation of FID with precision, recall, and $F_1$. We observe that the proposed measure is particularly suitable for detecting a loss in recall.}
  \label{fig:f1_vs_budget_wide_recall_3}
\end{figure*}

\end{document}